\documentclass[sigconf, natbib=false]{acmart}
\AtBeginDocument{%
  }

\copyrightyear{2025}
\acmYear{2025}
\setcopyright{cc}
\setcctype{by-nc-sa}
\acmConference[CIKM '25]{Proceedings of the 34th ACM International Conference on Information and Knowledge Management}{November 10--14, 2025}{Seoul, Republic of Korea}
\acmBooktitle{Proceedings of the 34th ACM International Conference on Information and Knowledge Management (CIKM '25), November 10--14, 2025, Seoul, Republic of Korea}\acmDOI{10.1145/3746252.3761128}
\acmISBN{979-8-4007-2040-6/2025/11}

\RequirePackage[
  datamodel=acmdatamodel,
  style=acmnumeric,
  ]{biblatex}

\usepackage{algpseudocode}
\usepackage{algorithmicx}
\usepackage{algcompatible}
\usepackage{algorithm}

\usepackage{amsmath}

\usepackage{amssymb}
\usepackage{mathtools}
\usepackage{nicefrac}

\usepackage{microtype}
\usepackage{graphicx}
\usepackage{subfigure}
\usepackage{booktabs} 
\usepackage{multirow}
\usepackage{makecell}
\usepackage{array}
\usepackage{enumitem}
\usepackage{balance}

\newlist{enumerate*}{enumerate}{1}
\setlist[enumerate*]{label=(\roman*), nosep, afterlabel=\ }

\newcommand{\mypapertitle}{CUFL} 
\newcommand{\mypapertitlefull}{\textit{\textbf{C}urriculum guided personalized s\textbf{U}bgraph \textbf{F}ederated \textbf{L}earning}}

\newcommand*{\CMI}[1]{{\color{black}#1}}
\newcommand*{\MG}[1]{{\color{black}#1}}

\newtheorem{theorem}{Theorem}

\newtheorem{assumption}{Assumption}

\newlength{\origtextfloatsep}
\newlength{\origfloatsep}

\begin{document}

\title{Curriculum Guided Personalized Subgraph Federated Learning}


\author{Minku Kang}
\orcid{0009-0006-4927-6122}
\affiliation{%
  \institution{Sungkyunkwan University}
  \city{Suwon}
  \country{Republic of Korea}
}
\email{netisen3@skku.edu}

\author{Hogun Park}
\authornote{Corresponding author.}
\orcid{0000-0003-0576-5806}
\affiliation{%
  \institution{Sungkyunkwan University}
  \city{Suwon}
  \country{Republic of Korea}
}
\email{hogunpark@skku.edu}

\renewcommand{\shortauthors}{Kang et al.}

\begin{abstract}

Subgraph Federated Learning (FL) aims to train Graph Neural Networks (GNNs) across distributed private subgraphs, but it suffers from severe data heterogeneity.
To mitigate data heterogeneity, weighted model aggregation personalizes each local GNN by assigning larger weights to parameters from clients with similar subgraph characteristics inferred from their current model states.
However, the sparse and biased subgraphs often trigger rapid overfitting, causing the estimated client similarity matrix to stagnate or even collapse.
As a result, aggregation loses effectiveness as clients reinforce their own biases instead of exploiting diverse knowledge otherwise available.
To this end, we propose a novel personalized subgraph FL framework called \emph{\mypapertitlefull{}} (\mypapertitle{}).
On the client side, \mypapertitle{} adopts \emph{Curriculum Learning} (CL) that adaptively selects edges for training according to their reconstruction scores, exposing each GNN first to easier, generic cross-client substructures and only later to harder, client-specific ones.
This paced exposure prevents early overfitting to biased patterns and enables gradual personalization.
By regulating personalization, the curriculum also reshapes server aggregation from exchanging generic knowledge to propagating client-specific knowledge.
Further, \mypapertitle{} improves weighted aggregation by estimating client similarity using fine-grained structural indicators reconstructed on a random reference graph.
Extensive experiments on six benchmark datasets confirm that \mypapertitle{} achieves superior performance compared to relevant baselines.
\CMI{Code is available at \url{https://github.com/Kang-Min-Ku/CUFL.git}.}



\end{abstract}




\maketitle

\section{Introduction}

\begin{figure}[!t]
    \centering
    \includegraphics[width=0.95\columnwidth]{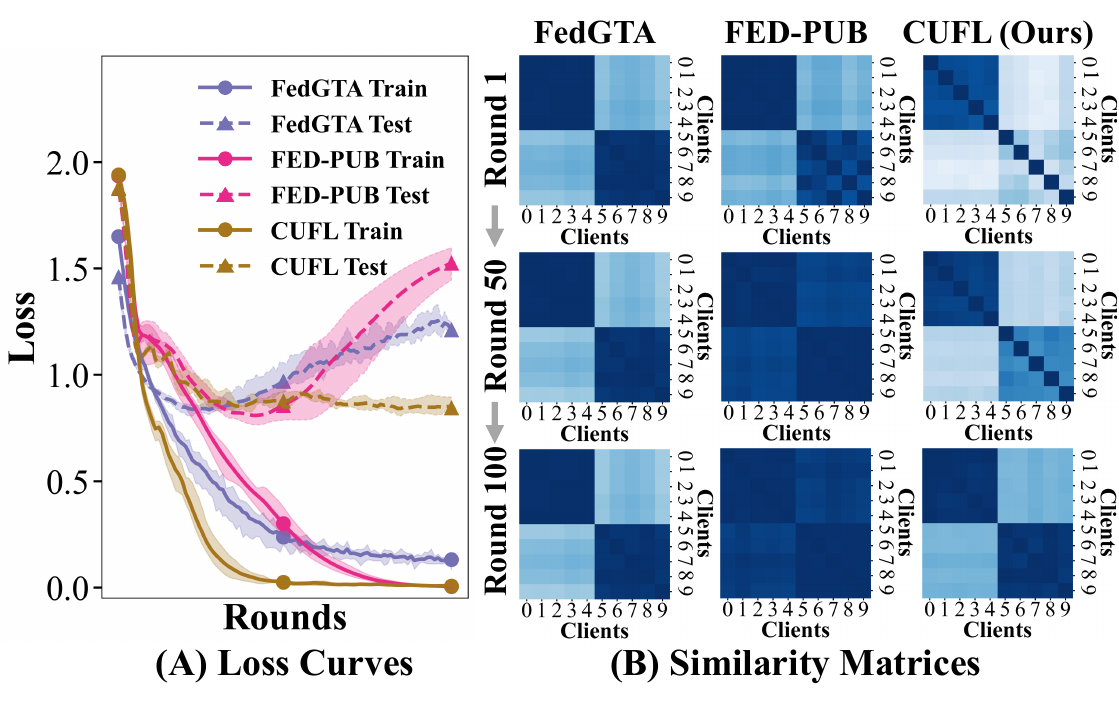}
    \vspace{-4mm}
    \caption{
     Training trends of three FL frameworks on Cora with 10 clients.
     (A) Cross-entropy loss curves imply that overfitting may occur rather frequently in personalized Subgraph FL.
     (B) The first five data-similar clients form one group, while the remaining five constitute another. 
     When overfitting sets in, client similarity matrices first plateau and can eventually crumble.
     CL prevents overfitting and, through its curriculum process, reshapes server aggregation step by step.
    }
    \Description{Intro Intuition}
    \label{fig:introMotivation}
    \vspace{-5mm}
\end{figure}


Graph Neural Networks (GNNs) \cite{DBLP:journals/tnn/ScarselliGTHM09} have achieved significant success on structured data \cite{DBLP:conf/www/Zhao0W24,park2025cimage,jung2025balancing}.
However, in practice, the global graph is fragmented across multiple clients, each holding a private subgraph that cannot be shared \cite{DBLP:conf/nips/TsoyK23}, thus limiting every client's view of the overall structure \cite{DBLP:conf/nips/ZhangYLSY21}.
Consequently, insufficient data leads to suboptimal GNN performance, highlighting the need for solutions to train performant GNNs without data disclosure.
In response to this issue, Subgraph Federated Learning (FL), a distributed GNN training framework, has been proposed \cite{liu2024federated}.
During each training round of Subgraph FL, clients train their local GNNs, and the server then aggregates the local GNN parameters into a global model.
However, due to heterogeneity in clients’ scarce and biased subgraphs, the server ends up averaging mismatched distributions, resulting in a global model that is suboptimal for many clients.
To mitigate this mismatch, weighted model aggregation assigns higher weights to clients with similar data distributions, tailoring server updates to each local GNN \cite{DBLP:conf/icml/YeNWCW23, DBLP:journals/corr/abs-2305-15706}.
Since data must stay private, client similarity---which is immediately transformed into aggregation weights---is inferred from current model states, thereby adjusting in step with the local fitting process.

Nevertheless, a key challenge persists: data sparsity and bias often induce rapid overfitting to suboptimal local patterns \cite{DBLP:journals/corr/abs-2401-02329, DBLP:conf/nips/LiLLJZ24}, pushing local GNNs to reinforce existing biases instead of absorbing the richer knowledge from weighted aggregation.
To investigate this challenge, we first split the global graph into two partitions with METIS \cite{karypis1997metis} and then repeatedly sample 50\% of the nodes from each partition, producing five partially overlapping subgraphs that share similar data properties.
FedGTA \cite{DBLP:journals/pvldb/LiWZZLW23} derives client similarity from label propagation features that combine soft labels with the structural context.
However, rapid overfitting (\figurename~\ref{fig:introMotivation}-(A)) drives similarity scores to extremes, causing the similarity matrix to lock in prematurely (\figurename~\ref{fig:introMotivation}-(B), left).
With the similarity matrix frozen, subsequent updates are confined to the initial ``most-similar'' subgroup, causing the training process to reinforce patterns already shared inside that group.
In FED-PUB \cite{DBLP:conf/icml/BaekJJYH23}, client similarity is measured coarsely by comparing average node embeddings that each local GNN outputs on a shared random graph.
FED-PUB likewise succumbs to collaboration lock-in due to rapid overfitting (\figurename~\ref{fig:introMotivation}-(A)).
The coarse functional nature of FED-PUB's similarity estimation produces an additional failure mode whereby the matrix shrinks over successive rounds and ultimately collapses~(\figurename~\ref{fig:introMotivation}-(B), middle).
Taken together, rapid overfitting anchors weighted aggregation in narrow local patterns, leaving little room for further enrichment on the server.
Accordingly, an adaptive mechanism that regulates the degree of local model fitting is essential for breaking the lock-in and unlocking the full benefits of server aggregation.

Motivated by this goal, we integrate \emph{Curriculum Learning} (CL) \cite{DBLP:conf/icml/BengioLCW09,DBLP:conf/icml/HacohenW19,DBLP:conf/iccv/KongLWT21,DBLP:conf/icml/WeinshallCA18}, which organizes training samples from ``easy'' to ``hard'', whereby easier samples remain close to data distribution, whereas harder ones embody client-specific characteristics \cite{DBLP:conf/iccv/VahidianKB0K0S023}.
Recognizing these insights, we propose a novel personalized Subgraph FL framework named {\mypapertitlefull} (\mypapertitle), which guides gradual personalization through a curriculum and supports it with a precise weighted aggregation method.
Our CL strategy incrementally incorporates graph edges based on continuously updated difficulty, evaluated by how well the local GNNs can reconstruct these edges.
Owing to their adaptivity to divergent server updates, automatic strategies furnish an appropriate curriculum for each client without additional communication or global statistics.
With this design in place, each local GNN traverses a staged personalization trajectory---initially fitting to cross-client regularities in the early rounds and then gradually shifting its focus to client-specific details in later rounds.
This staged fitting prevents each local GNN from overfitting to its own biased patterns too early.
Beyond this, CL also reshapes the dynamics of server aggregation. 
Because weighted aggregation adapts to evolving local model states, the knowledge propagated by the server shifts from broadly shared patterns in the early rounds to increasingly client-specific information as training progresses~(\figurename~\ref{fig:introMotivation}-(B), right).
\CMI{However, sustaining the benefits of this evolving flow requires high-resolution estimate of client similarity.
To this end, under the data constraints of FL \cite{DBLP:conf/nips/TsoyK23}, client similarity is measured through comparisons of reconstructed graph structures on a shared random reference graph.
Since models trained on analogous data produce similar node embedding distributions for identical inputs \cite{DBLP:conf/icml/Kornblith0LH19}, their reconstructions of the reference graph are likewise comparable.
These reconstructions provide fine-grained structural indicators without disclosing client data, thereby enabling high-resolution and privacy-preserving similarity estimation.}
Comprehensive experiments on six benchmark datasets demonstrate outstanding performance of \mypapertitle{}.
Further analyses show the impact of CL on local GNN personalization as well as the robustness of \mypapertitle{}’s similarity estimation method.

\vspace{-2mm}
\section{Related Works}

\subsection{Subgraph Federated Learning}

Federated Learning (FL) \cite{DBLP:journals/mlc/WenZLCCZ23} often struggles in subgraph tasks by overlooking structural characteristics, motivating subgraph FL frameworks for the distributed subgraph setting \cite{liu2024federated}.


Several frameworks transfer knowledge between global and local models.
FedSpray \cite{DBLP:conf/kdd/FuCZ0L24} distills unbiased soft labels from a global feature-structure encoder to local models.
FedGF \cite{DBLP:conf/wsdm/Zhou00LYFLWZW25} employs graph atoms with personalized weights to inject customized global structural knowledge into local models.
While these frameworks enhance local models with global information, \mypapertitle{} improves local GNNs without global guidance.
In the opposite direction, FedTAD \cite{DBLP:conf/ijcai/ZhuLWWHL24} uploads class-specific knowledge from local GNNs to the global model according to each client’s reliability for that class.

Alternatively, another line of research focuses on expanding local subgraphs.
FedGNN \cite{DBLP:journals/corr/abs-2102-04925} extends overlapping nodes from other subgraphs.
In a related manner, FedSage+ \cite{DBLP:conf/nips/ZhangYLSY21} generates missing neighbors and mends the local subgraph by incorporating them.
However, these subgraph augmentation methods suffer from unintended data exposure and communication overhead.

Weighted model aggregation, which tailors collaboration to each client, has also been studied.
\CMI{FedSG \cite{DBLP:journals/tetci/WangGQLL24} measures client similarity by separately comparing the parameters of each component in the local model, which is computationally inefficient.}
FedGTA \cite{DBLP:journals/pvldb/LiWZZLW23} groups clients from mixed moments of neighbor features obtained via label propagation.
However, rapid overfitting locks server aggregation into peer groups, curbing the intake of diverse, beneficial knowledge.
In a separate line of work, FED-PUB \cite{DBLP:conf/icml/BaekJJYH23} points out data heterogeneity induced by community structure and proposes a solution by evaluating the similarity of functional embeddings on a random graph to predict communities.
Under rapid overfitting, client similarity estimation of FED-PUB gradually collapses as training progresses.


\begin{figure*}[!htbp]
    \centering
    \includegraphics[width=0.95\textwidth]{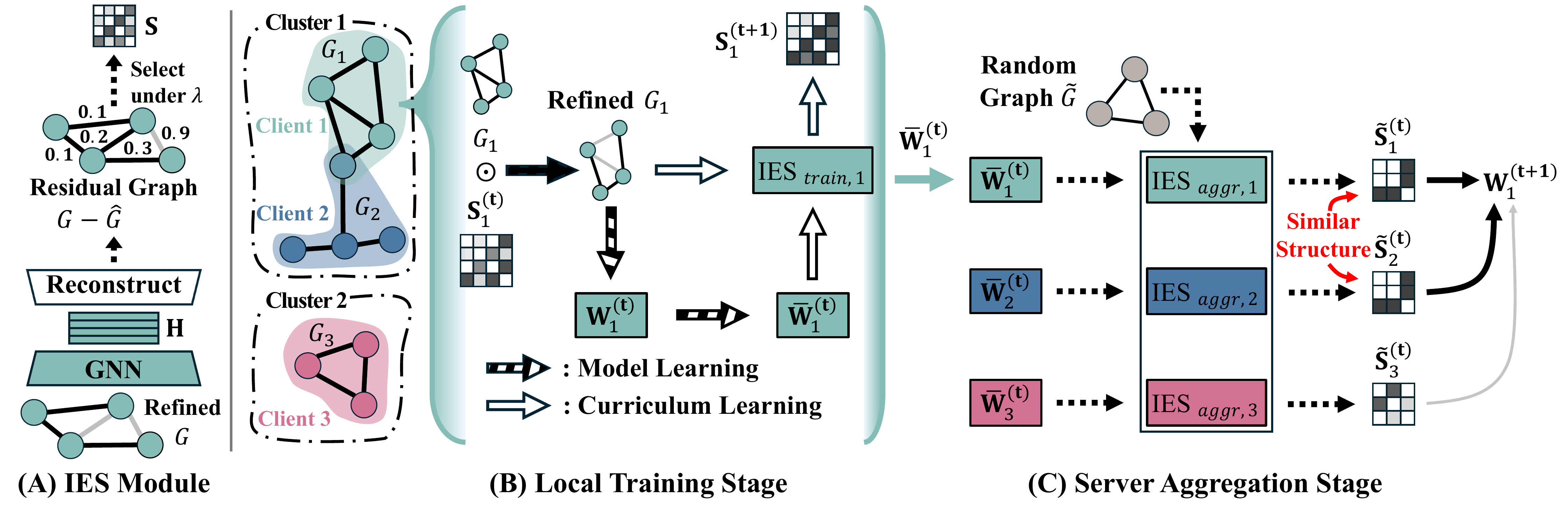}
    \vspace{-4mm}
    \caption{Overview of the \mypapertitle{} framework for Client 1. \textbf{(A)} \textbf{Incremental Edge Selection (IES)} module to determine ``well-expected'' edges.
    \textbf{(B)} \textbf{Local Training Stage} trains the local GNN and the subgraph mask.
    \textbf{(C)} \textbf{Server Aggregation Stage} performs personalized aggregation.
    Clients 1 and 2 share more information with each other than with Client 3, given their similar data.
    }
    \Description{Main Figure}
    \label{fig:method}
    \vspace{-0.1in}
\end{figure*}

\vspace{-3mm}
\subsection{Curriculum Graph Learning}

Curriculum Learning (CL) comprises three components.
A score function assigns difficulty, a scheduling function orders training samples, and a pacing function controls the pace of data presentation.
CL strategies are categorized by whether the learning order changes continuously during training.
Pre-defined strategies use fixed learning orders, while automatic strategies adaptively adjust them during training by accepting feedback from the model \cite{DBLP:conf/ijcai/LiW023}.

Among CL strategies designed for graph data, CLNode \cite{DBLP:conf/wsdm/WeiGZ00H23} utilizes a pre-trained GNN to pre-define node difficulty, considering both node labels and features.
On the other hand, RCL \cite{DBLP:conf/nips/ZhangWZ23} integrates confident edges incrementally based on the current model expectations.
In addition to strategies, the effectiveness of CL has been widely studied both empirically and theoretically.
\cite{DBLP:conf/icml/BengioLCW09,DBLP:conf/iccv/KongLWT21,DBLP:conf/icml/WeinshallCA18, DBLP:conf/icml/HacohenW19} reveal that CL contributes to better generalization, robust optimization, and faster convergence.
Furthermore, \cite{DBLP:conf/icml/HacohenW19} proves its ability to subtly steepen the optimization landscape without altering the global optimum.
Following these insights, \cite{DBLP:conf/iccv/VahidianKB0K0S023} demonstrates that CL alleviates data heterogeneity in FL.
The study finds that CL is particularly useful during early training rounds, as the ``easier'' samples tend to be closer in distribution.
While their work illustrates the relationship between CL and FL, it is limited to centralized FL.

\vspace{-2mm}
\section{Preliminaries}

Given $K$ clients, each client $k$ owns a local subgraph $G_k=(\mathcal{V}_k, \mathcal{E}_k)$, which constitutes a substructure of the global graph $G=(\mathcal{V}, \mathcal{E})$, where $\mathcal{V}_k \subseteq \mathcal{V}$ and $\mathcal{E}_k \subseteq \mathcal{E}$.
$\mathcal{V}_k$ refers to a set of $|\mathcal{V}_k|$ nodes, and $\mathcal{E}_k$ to a set of $|\mathcal{E}_k|$ edges.
Client $k$'s $i$-th node $v_{k,i}\in\mathcal{V}_k$ is represented by a feature vector $\boldsymbol{x}_{k,i}\in\mathbb{R}^{d_{\boldsymbol{x}}}$, where $d_{\boldsymbol{x}}$ indicates its dimensionality.
The edges in $G_k$ are encoded in its adjacency matrix $\mathbf{A}_k\in\{0,1\}^{|\mathcal{V}_k| \times |\mathcal{V}_k|}$.
Specifically, $\mathrm{A}_k[i,j]=1$ if a link exists between the $i$-th and $j$-th nodes in $\mathcal{V}_k$, otherwise $\mathrm{A}_k[i,j]=0$.

\subsection{Graph Neural Networks}

Graph Neural Networks (GNNs) typically follow a message-passing framework \cite{DBLP:conf/icml/GilmerSRVD17}, iteratively aggregating and combining node representations to refine each node.
GNNs express an $i$-th node $v_i$ at the $l$-th layer as follows (the client index $k$ is omitted for clarity):
\begin{equation}
    \boldsymbol{h}_i^{l+1} = \text{COMB}^{l}\big(\boldsymbol{h}_i^{l}, \text{AGG}^{l}(\{\boldsymbol{h}_j^{l} : v_j \in \mathcal{N}(v_i)\})\big),
    \label{2-2:eq:gnn}
\end{equation}
where $\boldsymbol{h}_i^{l}$ denotes the representation of $v_i$ at the $l$-th layer.
The representation is initialized as $\boldsymbol{h}_i^{0}=\boldsymbol{x}_i$, and the final output is described as $\boldsymbol{h}_i$.
$\mathcal{N}(v_i)$ indicates the set of neighbors of $v_i$.
$\text{AGG}^l$ aggregates representations of $\mathcal{N}(v_i)$, and $\text{COMB}^l$ combines the previous representation of $v_i$ with the aggregated representation.

\subsection{Personalized Subgraph FL Optimization}


In Subgraph Federated Learning (FL), data heterogeneity hinders the effectiveness of a one-size-fits-all model, which often yields suboptimal performance for some clients.
In contrast, personalizing GNNs to each client's objective mitigates this heterogeneity.
The weighted model aggregation facilitates personalization by promoting parameter sharing among mutually beneficial peers.
The resulting personalized Subgraph FL objective \cite{DBLP:conf/icml/BaekJJYH23} is formulated as:
\begin{equation}
\begin{split}
    & \min_{\{\mathbf{W}_k\}_{k=1}^K}\sum_{G_k \subseteq G} \mathcal{L}(G_k; \mathbf{W}_k), \mathbf{W}_k \leftarrow \sum_{n=1}^{K} \alpha_{kn} \cdot \bar{\mathbf{W}}_n \hfill \\
    & \quad\quad\quad\quad \text{with } \alpha_{kp} \gg \alpha_{kq} \text{ for } G_p \subseteq C \text{ and } G_q \not\subseteq C,
    \label{2-2:eq:psfl}
\end{split}
\end{equation}
where $\mathcal{L}$ denotes the task-specific loss, $\mathbf{W}_k$ the aggregated parameters for client $k$, $\bar{\mathbf{W}}_k$ the locally trained parameters for client $k$, and $C$ a cluster of clients with similar data distributions.
$\alpha_{kn}$ is the coefficient for weighted aggregation between clients $k$ and $n$, which takes higher values for clients with subgraphs that are part of the same cluster.
\section{\mypapertitle{} Framework}

\subsection{Local Training Stage}

To enhance the benefits of weighted model aggregation, we employ CL, which flexibly tunes personalization throughout local training.

\subsubsection{Incremental Edge Selection (IES)
\label{section:IES}
}

We adopt Incremental Edge Selection (\emph{IES})~\cite{DBLP:conf/nips/ZhangWZ23} for dynamically generating personalized edge masks based on the understanding of the current model.
This automatic CL strategy fits well with personalized Subgraph FL, as it remains adaptable to divergent server updates without additional communication or global statistics.


The objective of the IES module for client $k$ is to optimize the learnable mask matrix $\mathbf{S}_k\in\mathbb{R}^{|\mathcal{V}_k| \times |\mathcal{V}_k|}$, which refines the local subgraph structure.
Specifically, at round $t$, the adjacency matrix of the input graph $G_k^{(t)}$ is derived as $\mathbf{S}_k^{(t)} \odot \mathbf{A}_k$, where $\odot$ indicates the Hadamard product, and $\mathbf{A}_k$ represents the adjacency matrix of the local subgraph $G_k$.
The difficulty of edges in IES is inversely proportional to the similarity between the current embeddings of their connected nodes.
As illustrated in Figure \ref{fig:method}-(A), the node embedding matrix $\mathbf{H}_k^{(t)}$, whose $i$-th row $\boldsymbol{h}_{k,i}^{(t)}$ corresponds to the output embedding of the $i$-th node extracted from the local GNN using the refined local subgraph, is assembled.
Then, the reconstructed graph $\hat{G}_k^{(t)}$ is produced through $\mathbf{H}_k^{(t)}$.
In the adjacency matrix $\hat{\mathbf{A}}^{(t)}_k$ of $\hat{G}_k^{(t)}$, the edge weight between nodes $i$ and $j$ is defined as $\kappa(\mathbf{H}^{(t)}_k[i,:],\mathbf{H}^{(t)}_k[j,:])$, where $\kappa$ is a kernel function implemented as cosine similarity.
IES prioritizes ``well-expected'' edges with higher weights of $\hat{G}_k^{(t)}$ for inclusion in training.

IES controls the pace of local training with an aging parameter $\lambda^{(t)}$.
That is, $\lambda^{(t)}$ quantitatively determines which edge qualifies as a ``well-expected'' edge.
$\lambda^{(t)}$ is defined as $g_\lambda(t)=\min(\frac{\zeta}{R} t,1)$, where $R$ is the total number of rounds and $\zeta$ designates the round when the full local subgraph is introduced.
The overall loss is denoted as:
\begin{equation}
\begin{split}
    \min_{\mathbf{S}_k} \, \sum_{i,j} \mathrm{S}_k[i,j] \left( \left\| \mathrm{A}_k[i,j] - \hat{\mathrm{A}}^{(t)}_k[i,j] \right\| - \lambda^{(t)} \right) \\
    + \frac{\gamma}{2} \left\| \mathbf{S}_k - \mathbf{S}_k^{(t)} \right\|,
    \label{4:eq:clloss}
\end{split}
\end{equation}
where $\gamma$ is a regularization coefficient and $\mathbf{S}_k^{(t)}$ indicates the current mask matrix.
The first term increases the mask weights for edges whose residual errors exceed $\lambda^{(t)}$.
Among them, easier edges with smaller residual errors show faster growth in mask weights, leading to their earlier inclusion in training.
The second term regularizes the mask to prevent abrupt changes, providing a smooth curriculum.
Hence, IES initiates training with easier substructures and then incrementally admits harder substructures.
At the outset of training, local GNNs predominantly fit low-frequency spectral components, which encode generic structural knowledge shared across clients \cite{DBLP:conf/icml/RahamanBADLHBC19, DBLP:conf/nips/TanWHY24, DBLP:journals/corr/abs-2502-13732}.
Consequently, early-round substructures exhibit highly similar distributions across clients, and as the curriculum progressively releases client-specific edges, each model personalizes gradually.
This staged fitting schedule suppresses rapid overfitting by guiding each model from shared patterns to individualized patterns.
Furthermore, by altering local training, the curriculum reconfigures server aggregation---initial rounds share generic knowledge, while subsequent rounds deliver more personalized updates.


\subsubsection{Local Model Optimization}

In the first round, all local GNNs are initialized with identical weights.
We then warm up client $k$'s subgraph mask $\mathbf{S}_k^{(1)}$ in $\text{IES}_{train,k}$ using a GNN pre-trained with FedProx \cite{DBLP:conf/mlsys/LiSZSTS20}.
This yields a cohesive initialization that keeps early-round mask updates aligned across clients \cite{DBLP:conf/iclr/NguyenWMSR23}.


During local training, the local GNN and the subgraph mask are optimized alternately (Figure \ref{fig:method}-(B)).
Algorithm \ref{4:alg} first applies the current mask $\mathbf{S}_k^{(t)}$ to the local subgraph $G_k$ (Line 3), and then trains the local GNN on the refined subgraph $G_k^{(t)}$ (Line 4).
Complementing CL, we add the proximal term \cite{DBLP:conf/mlsys/LiSZSTS20} to the training loss, further constraining the magnitude of local updates.
After training the local GNN, the latent node embedding matrix $\mathbf{H}_k^{(t)}$ is computed (Line 5), and the adjacency matrix is reconstructed using $\mathbf{H}_k^{(t)}$ (Line 6).
Next, $\mathbf{S}^{(t)}_k$ of $\text{IES}_{train,k}$ is optimized (Lines 7-8).
The $\operatorname{clip}(\cdot)$ function keeps all edge weights in the interval $[0,1]$.
Finally, the subgraph mask $\mathbf{S}_k^{(t+1)}$ and the aging parameter $\lambda^{(t+1)}$ are stored on the client for the next round (Line 10).

\subsection{Server Aggregation Stage}

The success of recent personalized FL frameworks \cite{DBLP:conf/icml/YeNWCW23, DBLP:journals/corr/abs-2305-15706} underscores the efficacy of weighted model aggregation, which lets each client receive favorable parameters from helpful peers while avoiding detrimental ones.
In this section, we present a method for deriving fine-grained signals that precisely estimate client similarity, and examine how CL reconfigures weighted model aggregation.



\setlength{\textfloatsep}{3mm}
\setlength{\floatsep}{3mm}

\begin{algorithm}[t]
\caption{Local Training Stage for Client $k$}
\begin{flushleft}
\textbf{Input}: Local subgraph $G_k$ whose adjacency matrix is $\textbf{A}_k$, current round $t$, mask matrix $\textbf{S}_k^{(t)}$, aggregated local GNN parameters $\textbf{W}_k^{(t)}$, age parameter $\lambda^{(t)}$, total number of epochs $E$, and regularization 
coefficients $\beta$ and $\gamma$      
\\
\textbf{Output}: Trained local GNN parameters $\Bar{\mathbf{W}}_k^{(t)}$, mask matrix $\mathbf{S}_k^{(t+1)}$, and age parameter $\lambda^{(t+1)}$    
\end{flushleft}
\begin{algorithmic}[1] 
\STATE $\bar{\mathbf{W}}_k^{(t)} \leftarrow \mathbf{W}_k^{(t)}$
\FOR{each local epoch $e$ from 1 to $E$}
    \STATE \parbox[t]{216pt}{Refine the local subgraph such that $G_k^{(t)}$ has the adjacency matrix $\mathbf{S}_k^{(t)} \odot \mathbf{A}_k$}
    \STATE $\bar{\mathbf{W}}_k^{(t)} \leftarrow \arg\min_{\mathbf{W}_k} \mathcal{L}(G_k^{(t)}; \mathbf{W}_k) 
    + \frac{\beta}{2}\|\mathbf{W}_k - \mathbf{W}_k^{(t)}\|$
    \STATE Get node embedding matrix $\mathbf{H}_k^{(t)}$ of $G_k^{(t)}$ with $\bar{\mathbf{W}}_k^{(t)}$
    \STATE \parbox[t]{216pt}{Compute reconstructed adjacency matrix $\hat{\mathbf{A}}_k^{(t)}$, where each edge weight is $\kappa(\mathbf{H}^{(t)}_k[i,:],\mathbf{H}^{(t)}_k[j,:])$}
    \STATE Optimize $\mathbf{S}^{(t)}_k$ according to Equation \eqref{4:eq:clloss}
    \STATE $\mathbf{S}^{(t)}_k \leftarrow \operatorname{clip}(\mathbf{S}^{(t)}_k)$
\ENDFOR
\STATE $\mathbf{S}_k^{(t+1)} \leftarrow \mathbf{S}_k^{(t)}$, and 
$\lambda^{(t+1)} \leftarrow g_\lambda(t+1)$
\end{algorithmic}
\label{4:alg}
\end{algorithm}

\subsubsection{Client Similarity Estimation via Node Embedding Distributions}

The intuition behind the proposed method is that local GNNs trained on subgraphs with similar properties exhibit comparable node embedding distributions for the same input \cite{DBLP:conf/icml/Kornblith0LH19}.
Such similarity in node embedding distributions arises as these local GNNs converge toward similar directions \cite{DBLP:conf/nips/CunKS90}.
Consider two data-similar clients $k$ and $n$.
For the same input, node embeddings close in the embedding space of $k$ remain close in the embedding space of $n$.
This consistency is also preserved when reconstructing the graph from node embeddings, resulting in reconstructed graphs of $k$ and $n$ having analogous structures.
Building on this, we feed a shared random graph to all clients.
After each reconstructs it with its local GNN, we compare the reconstructed structures to evaluate client similarity.
\CMI{By indirectly comparing embedding distributions through the reconstructed graphs, our similarity estimation approach operates at a fine-grained level.
At the same time, as the random graph is generated regardless of local subgraphs, the reconstruction-based comparison safeguards data privacy.}
The random graph $\Tilde{G}=(\Tilde{\mathcal{V}}, \Tilde{\mathcal{E}})$ is generated with a stochastic block model \cite{DBLP:journals/ans/LeeW19} because reconstructing multiple blocks offers richer structural cues than a single block.
Each node feature is initialized from a Gaussian distribution.
The random graph is reconstructed using its latent node embeddings from the local GNN, with edge weights computed based on the cosine similarity between embeddings of connected nodes.

To avoid redundant or noisy comparisons, we selectively compare the ``well-expected'' substructures, identified through the additional $\text{IES}_{aggr}$ module.
More precisely, we employ the subgraph mask $\Tilde{\mathbf{S}}_k$ of $\text{IES}_{aggr,k}$, optimized for the random graph via Equation~\eqref{4:eq:clloss}, as the indicator in client similarity estimation (Figure \ref{fig:method}-(C)).
This approach not only ensures comparisons on ``well-expected'' substructures but also preserves the continuity of client similarity estimation.
The similarity index between \MG{arbitrary client $k$ and $n$} at round $t$ is measured using Linear CKA \cite{DBLP:conf/icml/Kornblith0LH19}, expressed as:
\begin{equation}
    \operatorname{Sim}(\tilde{\mathbf{u}}_{k}^{(t)}\!,\!\tilde{\mathbf{u}}_{n}^{(t)})=
    \frac{\bigl\|\tilde{\mathbf{u}}_{k}^{(t)\top}\tilde{\mathbf{u}}_{n}^{(t)}\bigr\|_F^{2}}
         {\bigl\|\tilde{\mathbf{u}}_{k}^{(t)\top}\tilde{\mathbf{u}}_{k}^{(t)}\bigr\|_F\;
          \bigl\|\tilde{\mathbf{u}}_{n}^{(t)\top}\tilde{\mathbf{u}}_{n}^{(t)}\bigr\|_F},
    \quad
    \tilde{\mathbf{u}}_{k}^{(t)} = \operatorname{ext}\bigl(\tilde{\mathbf{S}}_{k}^{(t)}\bigr),
    \label{4:eq:refinedSimilarity}
\end{equation}
where $\Tilde{\mathbf{S}}_k^{(t)}\in\mathbb{R}^{|\Tilde{\mathcal{V}}| \times |\Tilde{\mathcal{V}}|}$ denotes the mask for the random graph of client $k$ at round $t$.
The $\operatorname{ext}(\cdot)$ function extracts all edge weights from the mask matrix and converts them into a $|\Tilde{\mathcal{E}}|$-dimensional vector, subsequently setting the lowest percentiles of values to zero.
This vectorization can improve computational efficiency, while weight pruning helps preserve clearer ``well-expected'' substructures.
The effectiveness of our client similarity index stems from its edge-wise multiplication, which assigns higher similarity to pairs with closely matching reconstructed random graph structures.

\setlength{\textfloatsep}{\origtextfloatsep}
\setlength{\floatsep}{\origfloatsep}
\begin{figure}[!t]
    \centering
    \includegraphics[width=0.90\columnwidth]{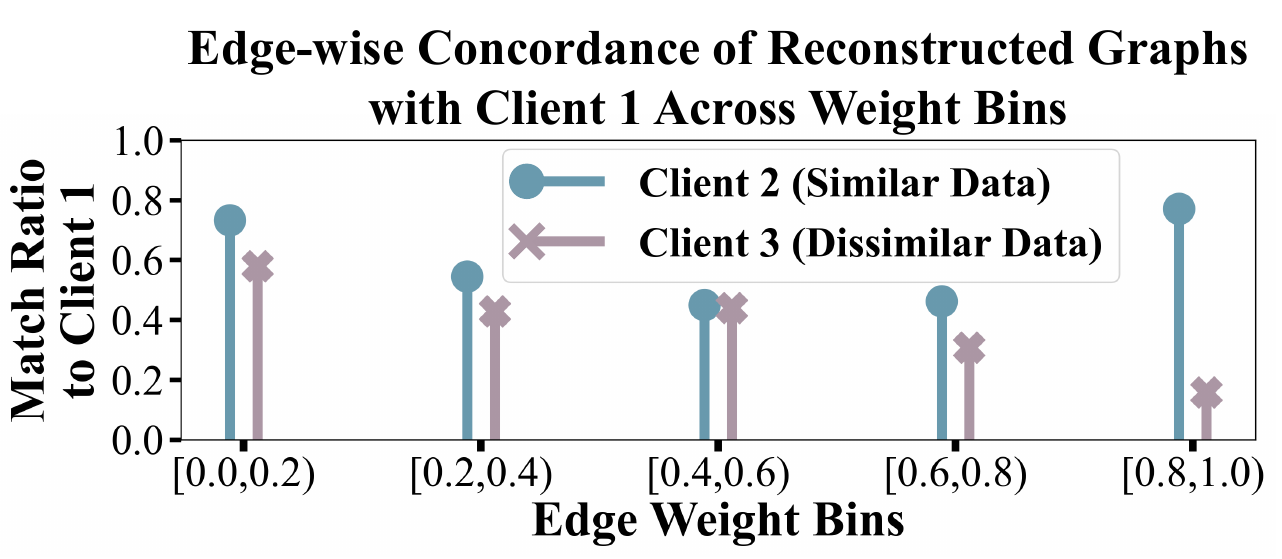}
    \vspace{-3mm}
    \caption{
    Edge-wise bin-match ratios with respect to Client 1 on the Cora dataset.
    The lollipop chart represents the ratio of identical edges from the reconstructed random graph that fall into the same bin as those in Client 1's graph.
    }
    \Description{Method Intuition}
    \label{fig:clNecessity}
    \vspace{-5mm}
\end{figure}

\paragraph{Toy Experiment.}
To illustrate how well such reconstructions capture client similarity, we split the Cora dataset into two partitions using METIS \cite{karypis1997metis}, twice sample 50\% of nodes from the first to create overlapping subgraphs for Clients 1 and 2, and sample an equal-size subgraph from the second partition as Client 3.
Each client’s local GNN then reconstructs the same random graph by computing an adjacency matrix based on the cosine similarities among its learned node embeddings. 
Figure~\ref{fig:clNecessity} quantifies the structural alignment between reconstructed graphs by using Client 1 as reference.
Client 1's reconstructed edges are first grouped into five weight bins.
For each of Clients 2 and 3, a bin-match is counted whenever the corresponding edge falls into the same bin as in Client 1.
The match count in each bin is normalized by Client 1's edge total for that bin, so a higher ratio means stronger edge-wise agreement in that bin with the reference reconstruction.
Empirical results show that clients with similar data produce more closely matched reconstructed random graph structures than clients with dissimilar data.
Notably, data-homogeneous clients share a substantial portion of their ``well-expected'' edges.
Accordingly, we primarily focus on the ``well-expected'' substructures to refine client similarity estimation.

\subsubsection{Personalized Parameter Aggregation}
We now update local GNN parameters through weighted aggregation guided by client similarity.
The parameters are aggregated as follows:
\begin{equation}
    \mathbf{W}_k^{(t+1)} \leftarrow
    \sum_{n=1}^{K} \alpha_{kn}^{(t)} \,\bar{\mathbf{W}}_n^{(t)},
    \alpha_{kn}^{(t)} =
    \frac{\exp\bigl(\tau\operatorname{Sim}(\tilde{\mathbf{u}}_{k}^{(t)}\!,\! \tilde{\mathbf{u}}_{n}^{(t)})\bigr)}
         {\sum_{p=1}^{K} \exp\bigl(\tau\operatorname{Sim}(\tilde{\mathbf{u}}_{k}^{(t)}\!,\! \tilde{\mathbf{u}}_{p}^{(t)})\bigr)},
    \label{4-1:eq:aggregation}
\end{equation}
where $\mathbf{W}_k^{(t+1)}$ represents the aggregated model parameters, $\Bar{\mathbf{W}}_n^{(t)}$ denotes the locally trained model parameters, $\alpha_{kn}^{(t)}$ is the normalized similarity between clients $k$ and $n$, and $\tau$ is a scaling factor determining the intensity of collaboration.
This personalized aggregation encourages cooperation among clients sharing similar data, whose local GNNs behave in a similar manner.
In \mypapertitle{}, the aggregation procedure is further recalibrated through CL.
In the initial rounds, parameter sharing is broad because the curriculum begins with ``easier'' samples drawn from closely aligned data distributions.
As each local GNN becomes more personalized, the aggregation assigns relatively smaller weights to dissimilar clients while proportionally increasing the weight of data-aligned clients.

\subsection{Complexity Analysis}

In this section, we provide a complexity analysis of our proposed \mypapertitle{}.
In the local training stage, the main additional cost arises from the incorporation of CL.
The computational cost of learning the subgraph mask is $\mathcal{O}(|\mathcal{E}_k|\cdot|h_{k,i}|+|\mathcal{E}_k|)$, where $\mathcal{E}_k$ is the edge set of the local subgraph and $h_{k,i}$ denotes the output embedding of the $i$-th node produced by the local GNN.
Since the complexity of optimizing the subgraph mask is comparable to or even lower than that of the local GNN training \cite{DBLP:journals/corr/abs-1901-00596}, it introduces only a marginal computational burden.
In the server aggregation stage, constructing a fine-grained client indicator requires $\mathcal{O}(|\tilde{\mathcal{E}}|\cdot|h_{k,i}|+|\tilde{\mathcal{E}}|)$, where $\tilde{\mathcal{E}}$ is the edge set of the random graph.
As for communication overhead, each client transmits the mask of the random graph to the server, incurring a cost of $\mathcal{O}(|\tilde{\mathcal{E}}|)$.
\CMI{In \mypapertitle{}, the overall cost of server aggregation depends on the size of the random graph $\tilde{G}$, which is generally much smaller than the local subgraph.
Such independence from the local subgraph size \cite{DBLP:journals/pvldb/LiWZZLW23} and the model parameter size \cite{DBLP:journals/tetci/WangGQLL24} ensures that \mypapertitle{} remains scalable.}

\subsection{Theoretical Analysis}

In this section, we present a theoretical analysis that guarantees data-similar clients retain similar adjacency reconstructions for the same random graph, while remaining distinct from data-dissimilar clients.
This statement relies on standard assumptions of Lipschitz continuity for the reconstructability function and bounded embedding divergence for data-similar clients.

\begin{theorem}[Cluster Preservation]
\label{thm:community}
Suppose clients $k$ and $m$ belong to the same cluster $C$ of similar data properties, whereas client $n$ does not.
Let $\tilde{\mathbf{A}}_k^{(t)}$, $\tilde{\mathbf{A}}_m^{(t)}$, and $\tilde{\mathbf{A}}_n^{(t)}$ be their reconstructed adjacency matrices at round $t$ for the random graph.
Under the Lipschitz property of $\kappa$ and the coherence assumption of Equation \eqref{eq:coherence} in Appendix~\ref{sec:theory}, there exists $\xi>0$ such that
\begin{equation}
\label{eq:community-preserve}
    \|\tilde{\mathbf{A}}_k^{(t)} 
      - \tilde{\mathbf{A}}_m^{(t)}\|_F
    \;\;\le\;\;
    \frac{1}{\xi}
    \,\bigl\|\tilde{\mathbf{A}}_k^{(t)}
            - \tilde{\mathbf{A}}_n^{(t)}\bigr\|_F
    \;+\;
    \mathcal{O}\!\bigl(\epsilon_C\bigr),
\end{equation}
for some small $\epsilon_C>0$. Hence, same-cluster reconstructions remain closer than cross-cluster reconstructions.
\end{theorem}

The theorem implies that, with high probability, data-similar clients generate closely matching reconstructions of a (potentially global) random graph, whereas this correspondence fails for clients whose data differ.
Such a gap ensures that our method—when measuring similarity across clients—can effectively identify helpful peers and preserve their structure over the course of training.
The full proof and a more detailed discussion appear in Appendix~\ref{sec:theory}.
            

\begin{table*}[t]
\caption{
Node classification performance on FL frameworks over three different numbers of participating clients. The data is configured to have no overlapping edges between clients. The presented outcome is the mean and standard deviation of accuracy at the final round of training. The best result is \textbf{bold}, and the second result is \underline{underlined}.
}
\vspace{-4mm}
\begin{center}
\begin{small}
\resizebox{\textwidth}{!}{
\begin{tabular}{l ccc ccc ccc}
 \noalign{\vskip 0.3mm} \Xhline{2\arrayrulewidth} \noalign{\vskip 0.3mm}

\noalign{\vskip 0.3mm}
\multicolumn{1}{c}{} & \multicolumn{3}{c}{\textbf{Cora}} & \multicolumn{3}{c}{\textbf{CiteSeer}} & \multicolumn{3}{c}{\textbf{PubMed}} \\ 
\cmidrule(lr){2-4}\cmidrule(lr){5-7}\cmidrule(lr){8-10}
\multicolumn{1}{c}{\textbf{Frameworks}}                     & \textbf{5 Clients} & \textbf{10 Clients}     & \textbf{20 Clients}     & \textbf{5 Clients} & \textbf{10 Clients}     & \textbf{20 Clients}     & \textbf{5 Clients} & \textbf{10 Clients}     & \textbf{20 Clients}     \\ \noalign{\vskip 0.3mm} \hline
\noalign{\vskip 0.3mm}
\multicolumn{1}{l|}{Local}           & 80.70$\pm$0.63     & 80.10$\pm$0.23          & 77.81$\pm$1.25          & 67.38$\pm$0.29     & 71.26$\pm$0.52          & 70.40$\pm$1.05          
& 83.25$\pm$0.17     & 81.26$\pm$0.21          & 82.68$\pm$0.53          \\ \noalign{\vskip 0.3mm} \hline
\noalign{\vskip 0.3mm}
\multicolumn{1}{l|}{FedAvg}          & 81.37$\pm$0.43     & 75.57$\pm$1.51          & 74.37$\pm$0.81          & 70.69$\pm$0.53     & 66.19$\pm$1.73          & 68.80$\pm$1.48          & 85.60$\pm$0.14     & 81.67$\pm$0.39          & 82.33$\pm$0.34          \\
\multicolumn{1}{l|}{FedProx}         & 81.13$\pm$0.31     & 74.12$\pm$1.87          & 71.37$\pm$4.23          & 70.91$\pm$0.78     & 66.22$\pm$1.71          & 68.90$\pm$0.98          & 85.59$\pm$0.13     & 81.23$\pm$0.60          & 82.22$\pm$0.36          \\
\multicolumn{1}{l|}{FedAvgCL}          & \underline{81.97$\pm$0.75}                         & 80.52$\pm$0.38                          & 74.88$\pm$0.84                          & \underline{71.36$\pm$0.65}                         & 66.65$\pm$1.07                          & 71.72$\pm$1.62                          & 85.98$\pm$1.04                          & 85.08$\pm$0.41                          & 85.49$\pm$0.46                          \\
\multicolumn{1}{l|}{FedPer}          & 81.75$\pm$0.41     & 80.60$\pm$0.05          & 76.64$\pm$1.71          & 70.38$\pm$0.55     & 70.89$\pm$0.62          & 69.49$\pm$1.02          & 85.51$\pm$0.11     & 84.89$\pm$0.39          & 82.88$\pm$0.15          \\ 
\noalign{\vskip 0.3mm} \hline
\noalign{\vskip 0.3mm}
\multicolumn{1}{l|}{FedTAD}        & 78.29$\pm$0.45     & 77.04$\pm$0.60          & 77.75$\pm$0.34          & 69.32$\pm$0.43     & 70.10$\pm$0.60          & 67.18$\pm$0.47          & 84.79$\pm$0.32     & 84.27$\pm$0.31          & 83.45$\pm$1.61          \\
\multicolumn{1}{l|}{FedSpray}        & 75.64$\pm$0.92     & 74.08$\pm$0.52          & \underline{79.21$\pm$0.27}          & 71.27$\pm$0.42     & \underline{75.62$\pm$0.38}          & 72.24$\pm$0.55          & 82.92$\pm$0.46     & 84.12$\pm$0.13          & 84.58$\pm$0.20          \\
\multicolumn{1}{l|}{FedGNN}          & 81.31$\pm$0.38     & 71.93$\pm$0.86          & 75.48$\pm$1.12          & 70.72$\pm$0.18     & 63.68$\pm$1.42          & 66.52$\pm$1.30          & 83.84$\pm$0.09     & 77.24$\pm$0.43          & 81.15$\pm$1.37          \\
\multicolumn{1}{l|}{FedSage+}        & 78.99$\pm$0.96     & 77.60$\pm$2.60          & 77.27$\pm$4.58          & 70.01$\pm$0.61     & 68.86$\pm$2.25          & 65.36$\pm$3.93          & 84.38$\pm$0.16     & 84.49$\pm$1.01          & 80.87$\pm$1.30          \\
\multicolumn{1}{l|}{FedGTA}          & 81.74$\pm$0.37    & \underline{82.60$\pm$0.41}    & 78.68$\pm$0.55    & 71.24$\pm$0.31    & 74.96$\pm$0.32    & \underline{72.33$\pm$0.54}    & \textbf{87.36$\pm$0.07}    & \textbf{86.64$\pm$0.10} & \underline{85.88$\pm$0.20}    \\
\multicolumn{1}{l|}{FED-PUB}         & 81.72$\pm$0.16     & 81.79$\pm$0.11          & 77.90$\pm$1.69          & 71.28$\pm$0.20     & 72.80$\pm$1.30          & 70.97$\pm$1.13          & 85.64$\pm$0.14     & 85.44$\pm$0.39          & 85.10$\pm$0.27          \\ \noalign{\vskip 0.3mm} \hline
\noalign{\vskip 0.3mm}
\multicolumn{1}{l|}{\mypapertitle{} (Ours)} & \textbf{83.66$\pm$0.19}     & \textbf{83.93$\pm$0.25} & \textbf{79.35$\pm$0.38} & \textbf{72.04$\pm$0.19}     & \textbf{77.24$\pm$0.22} & \textbf{72.82$\pm$0.54} & \underline{86.54$\pm$0.06}     & \underline{86.22$\pm$0.05}    & \textbf{86.07$\pm$0.19} \\ \noalign{\vskip 0.3mm} \hline

\end{tabular}
}
\resizebox{\textwidth}{!}{
\begin{tabular}{c ccc ccc ccc}
 \noalign{\vskip 0.3mm} \hline \noalign{\vskip 0.3mm}

\multicolumn{1}{c}{} & \multicolumn{3}{c}{\textbf{Amazon-Computer}} & \multicolumn{3}{c}{\textbf{Amazon-Photo}} & \multicolumn{3}{c}{\textbf{ogbn-arxiv}} \\ 
\cmidrule(lr){2-4}\cmidrule(lr){5-7}\cmidrule(lr){8-10}
\multicolumn{1}{c}{\textbf{Frameworks}}                     & \textbf{5 Clients} & \textbf{10 Clients}     & \textbf{20 Clients}     & \textbf{5 Clients} & \textbf{10 Clients}     & \textbf{20 Clients}     & \textbf{5 Clients} & \textbf{10 Clients}     & \textbf{20 Clients}     \\ \noalign{\vskip 0.3mm} \hline
\noalign{\vskip 0.3mm}
\multicolumn{1}{l|}{Local}           & 88.96$\pm$0.40                         & 88.05$\pm$0.14                          & 86.69$\pm$0.48                          & 92.07$\pm$0.09                         & 91.63$\pm$0.24                          & 86.69$\pm$0.48                          & 68.49$\pm$0.13                          & 67.65$\pm$0.05                          & 68.60$\pm$0.68\\ \noalign{\vskip 0.3mm} \hline
\noalign{\vskip 0.3mm}
\multicolumn{1}{l|}{FedAvg}          & 85.69$\pm$0.93                         & 85.18$\pm$0.81                          & 82.77$\pm$1.09                          & 92.29$\pm$0.36                         & 86.82$\pm$1.24                          & 86.48$\pm$0.42                          & 68.68$\pm$0.13                          & 67.36$\pm$0.05                          & 68.03$\pm$0.93                          \\
\multicolumn{1}{l|}{FedProx}         & 86.37$\pm$0.61                         & 87.71$\pm$0.83                          & 82.40$\pm$1.43                          & 91.96$\pm$0.32                         & 87.77$\pm$0.72                          & 85.82$\pm$0.85                          & 68.50$\pm$0.11                          & 67.24$\pm$0.09                          & 67.92$\pm$0.96                          \\
\multicolumn{1}{l|}{FedAvgCL}          & 86.08$\pm$1.52                         & 88.06$\pm$0.36                          & 88.02$\pm$0.55                          & 92.55$\pm$0.13                         & 87.86$\pm$0.44                          & 87.35$\pm$0.53                          & 67.88$\pm$0.12                          & 67.64$\pm$0.39                          & 67.82$\pm$1.31                          \\
\multicolumn{1}{l|}{FedPer}          & 86.34$\pm$0.74                         & 87.76$\pm$0.62                          & 82.52$\pm$0.85                          & 91.63$\pm$0.52                         & 87.73$\pm$0.49                          & 86.84$\pm$0.44                          & 68.57$\pm$0.16                          & 67.47$\pm$0.28                    & 67.91$\pm$1.11          \\ 
\noalign{\vskip 0.3mm} \hline
\noalign{\vskip 0.3mm}
\multicolumn{1}{l|}{FedTAD}        & 83.04$\pm$0.67     & 86.69$\pm$1.10          & 83.55$\pm$1.51          & 91.01$\pm$0.34     & 87.97$\pm$1.01          & 85.97$\pm$1.31          & 68.33$\pm$0.33     & \underline{67.71$\pm$0.42}          & 68.34$\pm$0.13          \\
\multicolumn{1}{l|}{FedSpray}        & 89.31$\pm$0.49     & \underline{89.46$\pm$0.25}          & 87.98$\pm$0.52          & 92.59$\pm$0.07     & 91.15$\pm$0.50          & 88.07$\pm$0.21          & 64.77$\pm$0.19     & 62.74$\pm$0.08          & 62.32$\pm$0.06          \\
\multicolumn{1}{l|}{FedGNN}          & 87.86$\pm$0.41                         & 87.74$\pm$0.64                          & 82.78$\pm$0.85                          & 90.01$\pm$0.66                         & 90.35$\pm$0.42                          & 86.76$\pm$0.54                          & 67.16$\pm$0.37                          & 64.89$\pm$0.31                          & 67.71$\pm$0.98                          \\
\multicolumn{1}{l|}{FedSage+}        & 85.37$\pm$1.60                         & 83.62$\pm$1.41                          & 69.29$\pm$2.91                          & 90.47$\pm$0.71                         & 77.28$\pm$1.23                          & 77.72$\pm$1.45                          & 67.40$\pm$0.16                          & 66.11$\pm$0.11                          & 67.30$\pm$1.78                           \\
\multicolumn{1}{l|}{FedGTA}          & 86.45$\pm$0.30                         & 81.31$\pm$1.08                          & 84.62$\pm$0.18                          & 92.73$\pm$0.22                         & \underline{92.38$\pm$0.15}                    & 88.34$\pm$0.33                          & 65.68$\pm$0.10                          & 66.12$\pm$0.03                          & 64.51$\pm$0.70                           \\
\multicolumn{1}{l|}{FED-PUB}         & \underline{89.65$\pm$1.26}                         & 89.30$\pm$0.15                    & \textbf{88.96$\pm$0.24}                 & \underline{92.97$\pm$0.14}                         & 92.25$\pm$0.22                          & \underline{88.37$\pm$0.72}                    & \textbf{68.87$\pm$0.24}                          & 67.09$\pm$0.86                          & \underline{68.81$\pm$1.09}          \\ \noalign{\vskip 0.3mm} \hline
\noalign{\vskip 0.3mm}
\multicolumn{1}{l|}{\mypapertitle{} (Ours)} & \textbf{90.47$\pm$0.13}                         & \textbf{89.93$\pm$0.07}                 & \underline{88.78$\pm$0.14}                    & \textbf{93.01$\pm$0.07}                         & \textbf{92.47$\pm$0.07}                 & \textbf{88.95$\pm$0.19}                 & \underline{68.72$\pm$0.15}                          & \textbf{68.38$\pm$0.25}                 & \textbf{70.66$\pm$0.91} \\ \noalign{\vskip 0.3mm} \Xhline{2\arrayrulewidth}

\end{tabular}
}
\end{small}
\end{center}
\label{5:table:main}
\vspace{-5mm}
\end{table*}

\section{Experiments}

\subsection{Experimental Setups}

\subsubsection{Datasets
\label{expr:Data}}

We simulate a distributed subgraph environment using METIS \cite{karypis1997metis}, a non-overlapping partitioning algorithm that allows explicit control of the number of subgraphs for a well-structured experiments.
Experiments are conducted under the transductive setting, where nodes within each subgraph are divided into train, validation, and test sets in a 2:4:4 ratio.
To assess performance, We use six benchmark datasets: Cora, CiteSeer and PubMed, small citation graphs \cite{DBLP:conf/icml/YangCS16};
Amazon-Computer and Amazon-Photo, product graphs \cite{DBLP:journals/corr/abs-1811-05868};
ogbn-arxiv, a large citation graph \cite{DBLP:conf/nips/HuFZDRLCL20}.

\subsubsection{Baselines}

We compare \mypapertitle{} with \textbf{Local}, a baseline with local training only.
For general FL, we perform comparisons on \textbf{FedAvg} \cite{DBLP:conf/aistats/McMahanMRHA17}, a centralized FL baseline that performs weighted aggregation based on local dataset size;
\textbf{FedProx} \cite{DBLP:conf/mlsys/LiSZSTS20}, that adds a proximal term to FedAvg;
\textbf{FedAvgCL}, that applies the same CL strategy as \mypapertitle{} to FedAvg;
and \textbf{FedPer} \cite{DBLP:journals/corr/abs-1912-00818}, a personalized FL baseline that separates the model into shared and personalized layers.
For Subgraph FL, we employ \textbf{FedTAD} \cite{DBLP:conf/ijcai/ZhuLWWHL24} and \textbf{FedSpray} \cite{DBLP:conf/kdd/FuCZ0L24}, that leverage knowledge transfer;
\textbf{FedGNN} \cite{DBLP:journals/corr/abs-2102-04925} and \textbf{FedSage+} \cite{DBLP:conf/nips/ZhangYLSY21}, for data augmentation methods;
and \textbf{FedGTA} \cite{DBLP:journals/pvldb/LiWZZLW23} and \textbf{FED-PUB} \cite{DBLP:conf/icml/BaekJJYH23}, customizing collaboration for individual clients. 

\subsubsection{Hyperparameters}

The local learner is a two-layer Graph Convolution Network \cite{DBLP:conf/iclr/KipfW17} with a hidden dimension size of 128.
Models are trained for 100 rounds with 1 epoch on citation graphs, and for 200 rounds with 2 and 3 epochs on product graphs and ogbn-arxiv, respectively.
The local learning rate is chosen by grid search in \{0.01, 0.001\}, while $\text{IES}_{train}$ and $\text{IES}_{aggr}$ use rates of 0.0005 and 0.00001, respectively.
We set the coefficient for the proximal term $\beta$ and IES regularizer $\gamma$ to 0.001; adopt Adam for local optimization.
The pacing factor is $\zeta=1.5$, and in $ext$ lowest 30\% values are set to zero.
The scaling factor $\tau$ is selected via grid search over \{5, 10, adaptive $\tau$ scheduler\}, where the adaptive $\tau$ scheduler greedily adjusts each client’s $\tau$ for the next round using the performance of the previous round \cite{subramanian2023zeroth}.
We generate a random graph using the stochastic block model with five blocks of 100 nodes each, where node features are drawn from $\mathcal{N}(0,1)$.
Edges are added with probability 0.1 within partitions and none across partitions.

\vspace{-0.25cm}

\subsection{Main Results}

\subsubsection{Performance and Effectiveness}

Table \ref{5:table:main} reports node classification results in the node non-overlapping scenario, where clients inherently cluster based on similar data characteristics.
\mypapertitle{} consistently outperforms all baselines in nearly every setting.
Compared to FedTAD and FedSpray, \mypapertitle{} achieves higher average performance by at least (2.99\%, 3.50\%, 2.04\%) for 5, 10, and 20 clients, respectively.
\mypapertitle{} also surpasses augmentation methods such as FedGNN and FedSage+, yielding gains of at least (2.26\%, 6.70\%, 4.38\%).
The performance of data augmentation approaches drops as participants grow because missing edges proliferate.
By contrast, \mypapertitle{} scales gracefully as its curriculum guided personalization is executed entirely on the client.
Within the class of weighted model aggregation frameworks, \mypapertitle{} posts the best results, outperforming FedGTA and FED-PUB by minimum margins of (0.72\%, 1.58\%, 1.09\%).
FedGTA shows strong results when label propagation is favorable, as on PubMed with 5 clients.
However, it underperforms on datasets like ogbn-arxiv, where label propagation is less informative.
Unlike methods relying on dataset-specific assumptions, the server aggregation in \mypapertitle{} remains performant across datasets.

\begin{figure}[t]
    \centering
    \begin{small}
    \begin{center}
        \includegraphics[width=\columnwidth]{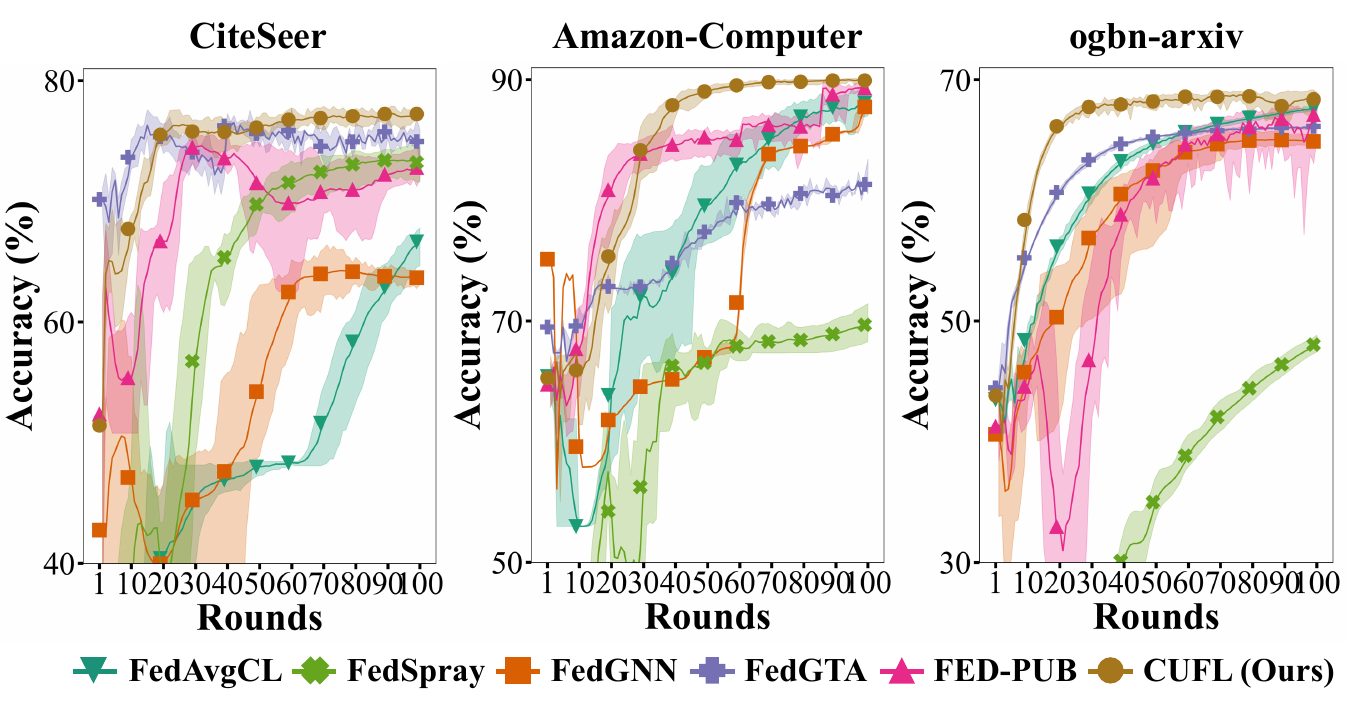}
        \vspace{-6mm}
        \caption{
        Accuracy curves for six FL frameworks.
        FedSpray training is stopped after 100 rounds.
        The plotted lines represent the average accuracy, and the shaded areas indicate the min-max range.
        }
        \label{fig:expr:learningCurve}
    \end{center}
    \end{small}
    \Description{Learning Curve}
    \vspace{-4mm}
\end{figure}

Alongside delivering exceptional performance, \mypapertitle{} demonstrates efficient learning. 
As Figure~\ref{fig:expr:learningCurve} indicates, the curriculum schedule lets \mypapertitle{} achieve fast convergence despite pronounced heterogeneity \cite{DBLP:conf/icml/HacohenW19}.
Moreover, the narrow performance spread in Figure~\ref{fig:expr:learningCurve} highlights the stability of the training process.
This stable behavior, attributed to the gradual personalization enabled by CL, results in relatively low standard deviations, as shown in Table \ref{5:table:main}.


\begin{figure}[t]
    \centering
    \includegraphics[width=0.85\columnwidth]{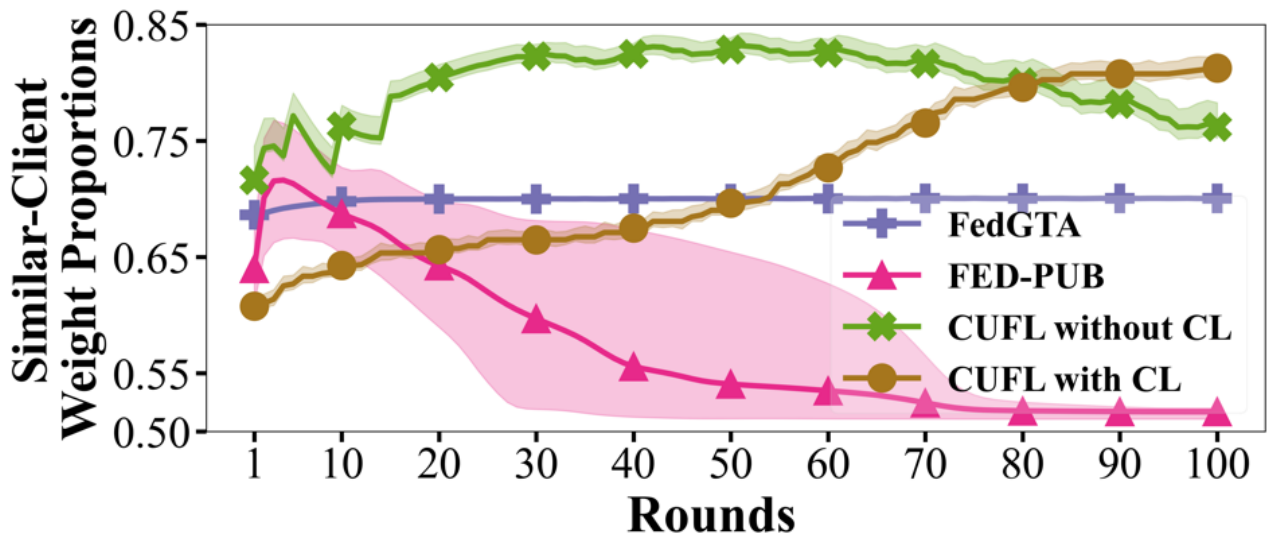}
    \vspace{-4mm}
    \caption{
    Evolution of weight proportions of data-similar clients during server aggregation on Cora with 10 clients under the node overlapping scenario.
    The shaded area shows the min-max range of the same community contributions.
    }
    \Description{Weight Curve}
    \label{fig:expr:weightCurve}
    \vspace{-5.5mm}
\end{figure}

\subsubsection{Impact of CL on Server Aggregation}

To examine how CL reshapes server aggregation process, we track the proportion of aggregation weight assigned to clients with similar data traits.
Figure \ref{fig:expr:weightCurve} plots this proportion, defined as $\tfrac{1}{K}\sum\limits_{k=1}^{K}\tfrac{\sum_{n \in C_k} \operatorname{Sim}(\tilde{\mathbf{u}}_{k}^{(t)},\tilde{\mathbf{u}}_{n}^{(t)})}{\sum_{n=1}^{K}\operatorname{Sim}(\tilde{\mathbf{u}}_{k}^{(t)},\tilde{\mathbf{u}}_{n}^{(t)})}$, where $K$ is the total number of clients, $C_k$ is the set of clients whose data properties are similar to client $k$, $\operatorname{Sim}(\cdot, \cdot)$ denotes inter-client similarity, and $\tilde{\mathbf{u}}_k^{(t)}$ is client $k$'s fine-grained information at round $t$.
FedGTA maintains the proportion almost constant.
Rapid overfitting drives the low variability in that proportion, ultimately making local GNNs reinforce biased patterns.
For both FED-PUB and \mypapertitle{} without CL, the proportion is initially high and even peaks in the early rounds.
As training continues, it drops while dissimilar clients gain more weight, eroding the intended collaboration.
In FED-PUB, the coarse functional similarity induces an abrupt decline in the proportion and amplifies its oscillations.
With CL enabled, the proportion begins modest, reflecting that local GNNs initially fit generic structural knowledge; as harder samples arrive, it rises smoothly, adapting the models to client-specific patterns.



\subsection{Ablation Studies}


\subsubsection{Regularization in Local Training Stage}

To prevent rapid overfitting at each client, \mypapertitle{} regularizes the local GNN with a proximal term \cite{DBLP:conf/mlsys/LiSZSTS20} that restrains local updates from diverging too far, and regularizes the local subgraph through CL.
Table \ref{table:divergence} shows that each mechanism alone raises accuracy and lowers variance, with CL providing the larger gain.
When combined, the proximal term and CL yield the most effective performance, confirming that their synergy is crucial for regulating the personalization of local GNNs.



\begin{table}[t]
\caption{Performance gain via Local Training Components}
\vspace{-2mm}
\begin{center}
\begin{small}
\resizebox{\columnwidth}{!}{
\begin{tabular}{ccccc}
\noalign{\vskip 0.3mm} \Xhline{2\arrayrulewidth} 
\hline
                                           & \multicolumn{2}{c}{\textbf{Cora}}                                                 & \multicolumn{2}{c}{\textbf{ogbn-arxiv}}                                           \\ \cline{2-5} 
\noalign{\vskip 0.3mm} \textbf{Conditions}                            & \multicolumn{1}{c}{\textbf{10 Clients}} & \multicolumn{1}{c}{\textbf{20 Clients}} & \multicolumn{1}{c}{\textbf{10 Clients}} & \multicolumn{1}{c}{\textbf{20 Clients}} \\ \hline
\noalign{\vskip 0.3mm}
\multicolumn{1}{l|}{None} & 81.55$\pm$0.33                          & 76.72$\pm$0.47                          & 67.56$\pm$0.24                          & 69.35$\pm$0.85                          \\
\multicolumn{1}{l|}{w/ Proximal Term}                & 82.07$\pm$0.31                          & 77.80$\pm$0.36                          & 67.83$\pm$0.20                          & 70.16$\pm$0.68                          \\
\multicolumn{1}{l|}{w/ CL}     & \underline{83.64$\pm$0.21}                          & \underline{79.05$\pm$0.28}                          & \underline{68.03$\pm$0.16}                          & \underline{70.55$\pm$0.53 }                         \\
 \noalign{\vskip 0.3mm} \hline \noalign{\vskip 0.3mm}
\multicolumn{1}{l|}
{w/ Proximal Term, CL (Ours)}       & \textbf{83.98$\pm$0.19}                          & \textbf{79.62$\pm$0.23}                          & \textbf{68.58$\pm$0.13}                          & \textbf{71.41$\pm$0.42}                          \\ \hline
\Xhline{2\arrayrulewidth}
\end{tabular}
}
\end{small}
\end{center}
\label{table:divergence}
\vspace{-0mm}
\end{table}

\begin{table}[t]
\caption{
Performance for combinations of CL strategy and ordering mechanism.
Each variant is written as \textit{CL strategy - ordering mechanism}.
The ordering mechanism is either a pre-trained evaluator or a random policy.
}
\vspace{-2mm}
\begin{center}
\begin{small}
\resizebox{\columnwidth}{!}{
\begin{tabular}{ccccc}
\noalign{\vskip 0.5mm} \Xhline{2\arrayrulewidth} 
\hline
                                           & \multicolumn{2}{c}{\textbf{PubMed}}                                                 & \multicolumn{2}{c}{\textbf{Amazon-Photo}}                                           \\ \cline{2-5} 
\noalign{\vskip 0.5mm} \textbf{CL strategies}                            & \multicolumn{1}{c}{\textbf{10 Clients}} & \multicolumn{1}{c}{\textbf{20 Clients}} & \multicolumn{1}{c}{\textbf{10 Clients}} & \multicolumn{1}{c}{\textbf{20 Clients}} \\ \hline
\noalign{\vskip 0.3mm}
\multicolumn{1}{l|}{CLNode - FedProx}                & 84.20$\pm$0.45                          & 83.18$\pm$1.03                          & 91.54$\pm$0.48                          & 86.96$\pm$0.54                          \\
\multicolumn{1}{l|}{IES - FedProx} & \underline{85.34$\pm$0.08}                          & \underline{85.59$\pm$0.23}                          & \underline{92.26$\pm$0.11}                          & \underline{88.65$\pm$0.49}                          \\ 
\multicolumn{1}{l|}{CLNode - Local}                & 80.06$\pm$1.80                          & 69.90$\pm$1.16                          & 90.57$\pm$0.64                          & 86.53$\pm$0.77                          \\
\multicolumn{1}{l|}{IES - Local}                & 85.01$\pm$0.17                          & 85.24$\pm$0.25                          & 90.61$\pm$0.14                          & 87.05$\pm$0.38                          \\
\multicolumn{1}{l|}{CLNode - Random}                & 79.98$\pm$0.34                          & 82.98$\pm$0.34                          & 90.18$\pm$0.05                          & 88.50$\pm$0.28                          \\
\multicolumn{1}{l|}{IES - Random}                & 78.12$\pm$0.09                          & 83.15$\pm$0.16                          & 89.65$\pm$0.51                          & 86.88$\pm$0.26                          \\
\noalign{\vskip 0.3mm} \hline \noalign{\vskip 0.3mm}
\multicolumn{1}{l|}
{IES}       & \textbf{86.22$\pm$0.02}                          & \textbf{86.17$\pm$0.18}                          & \textbf{92.53$\pm$0.04}                          & \textbf{89.05$\pm$0.18}                          \\ \hline
\Xhline{2\arrayrulewidth}
\end{tabular}
}
\end{small}
\end{center}
\label{table:cltype}
\vspace{-3mm}
\end{table}

\subsubsection{Effectiveness of Automatic CL Strategy}

\mypapertitle{} refines its curriculum on-the-fly from local GNN feedback.
To evaluate the effectiveness of this adaptive design, we evaluate \mypapertitle{} with two CL strategies, CLNode \cite{DBLP:conf/wsdm/WeiGZ00H23} and IES, and combine each strategy with three ordering policies: (i) a global model trained with FedProx, (ii) a local model trained only on the local subgraph (Local), and (iii) a uniformly random schedule (Random).
(i) and (ii) are pre-defined curricula; (iii) is order-agnostic.
As reported in Table \ref{table:cltype}, the pre-defined curricula perform noticeably worse in personalized Subgraph FL.
We attribute this gap to the widening mismatch between the evolving local GNNs and the fixed teacher models, which are difficult to keep sufficiently mature in a distributed subgraph environment.
Furthermore, the automatic curricula also surpass the random baseline.
The uniformly random schedule, being blind to model feedback, fails to achieve the gradual personalization essential to effective local GNN training.

\begin{figure}[t]
    \centering
    \includegraphics[width=0.95\columnwidth]{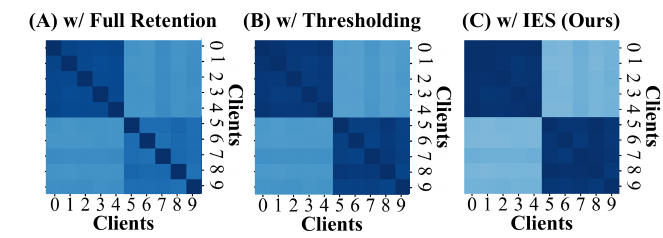}
    \vspace{-4mm}
    \caption{Heatmaps of estimated client similarity on Cora with 10 clients under the node overlapping scenario at the final round across various conditions.
    The first five clients form one group, and the other five form another.}
    \Description{CL Power on Community Detection}
    \label{fig:expr:clNecessity}
    \vspace{-4mm}
\end{figure}

\subsubsection{Edge Filtering Approaches for Client Similarity Estimation}

This part corroborates that ``well-expected'' edges in reconstructed graphs are appropriate for computing client similarity.
To this end, we compare similarity matrices under three settings---all edges, thresholding, and IES---in the node overlapping scenario.
Thresholding removes edges with weights below a preset threshold of 0.5.
As shown in Figure \ref{fig:expr:clNecessity}-(A), the contrast among estimated groups of clients with distinct data characteristics becomes blurred once all edges are retained.
To elaborate further, the output embeddings from the local GNN for inputs derived from a single distribution tend to be smoothed \cite{DBLP:journals/corr/abs-1911-08795,lee2024feature}.
This smoothing results in a reconstructed random graph with a significant portion of its edge weights at moderate values, causing substantial overlaps in the graph structures from most clients.
Thresholding can treat the problem of indistinct weighted model aggregation (Figure \ref{fig:expr:clNecessity}-(B)), but defining a suitable threshold remains challenging.
In contrast, optimizing IES enables pronounced weighted model aggregation without requiring comprehensive hyperparameter tuning (Figure \ref{fig:expr:clNecessity}-(C)).

\subsubsection{Varying Scaling Factor $\tau$
\label{expr:VSF}}


Because heterogeneity can change markedly according to data, the success of weighted model aggregation depends on considering both accurate client similarity estimation and the appropriate scaling of collaboration intensity.
Yet, while extensive attention has been devoted to estimating client similarity, the proper intensity of collaboration is unexplored.
Therefore, we investigate it by evaluating \mypapertitle{}'s performance across various scaling factors.
Alongside fixed $\tau$, we implement an adaptive $\tau$ scheduler that greedily adjusts each client's $\tau$ to its performance change relative to the previous round \cite{subramanian2023zeroth}.
As illustrated in Table \ref{5:table:tau}, a small $\tau$ (i.e., $\tau=3$) produces sub-optimal outcomes, causing insufficient cooperation among clients with similar data distributions.
The adaptive $\tau$ scheduler steadily delivers superior performance by providing each client with a tailored scaling factor.
When excluding this scheduler, the appropriate $\tau$ is 5 for 10 clients and 10 for 20 clients.
The trend implies that rising client counts exacerbate data heterogeneity, so directing communication toward the most relevant peers yields the greatest benefit.
However, when aggregation becomes overly inward-looking (i.e., $\tau=20$), performance paradoxically degrades, because of the absence of outward updates that could otherwise help navigate obstacles in the loss landscape.

\begin{table}[!ht]
\caption{Performance variation according to the scaling factor. ``adaptive'' refers to the adaptive $\tau$ scheduler.}
\vspace{-4mm}
\begin{center}
\begin{small}
\resizebox{\columnwidth}{!}{
\begin{tabular}{cllll}
\noalign{\vskip 0.3mm} \Xhline{2\arrayrulewidth} 
\hline
                                          & \multicolumn{2}{c}{\textbf{CiteSeer}}                                             & \multicolumn{2}{c}{\textbf{Amazon-Computer}}                                      \\ \cline{2-5} \noalign{\vskip 0.3mm}
\textbf{\textbf{Scaling Factors ($\tau$)}} & \multicolumn{1}{c}{\textbf{10 Clients}} & \multicolumn{1}{c}{\textbf{20 Clients}} & \multicolumn{1}{c}{\textbf{10 Clients}} & \multicolumn{1}{c}{\textbf{20 Clients}} \\ \hline
\noalign{\vskip 0.3mm}
\multicolumn{1}{c|}{3}                    & 76.76$\pm$0.72                          & 71.53$\pm$1.47                          & 89.72$\pm$0.08                          & 88.60$\pm$0.11                          \\
\multicolumn{1}{c|}{5}                    & \underline{77.17$\pm$0.88}                          & 71.93$\pm$1.11                          & \underline{89.83$\pm$0.13}                          & 88.72$\pm$0.15                          \\
\multicolumn{1}{c|}{10}                   & 76.37$\pm$0.41                          & \underline{72.17$\pm$0.96}                          & 89.61$\pm$0.09                          & \textbf{89.36$\pm$0.12}                          \\
\multicolumn{1}{c|}{20}                   & 75.78$\pm$0.49                          & 71.61$\pm$1.13                                        & 88.48$\pm$0.02                          & 87.58$\pm$0.08                                        \\ 
\multicolumn{1}{c|}{adaptive}             & \textbf{77.61$\pm$0.15}                          & \textbf{73.04$\pm$0.57}                          & \textbf{90.02$\pm$0.05}                          & \underline{88.90$\pm$0.05}                          \\ \hline 
\Xhline{2\arrayrulewidth}
\end{tabular}
}
\end{small}
\end{center}
\label{5:table:tau}
\vspace{-4.5mm}
\end{table}

\section{Conclusion}

In this work, we tackle the overlooked problem of rapid overfitting in weighted model aggregation.
Rapid overfitting constrains the benefits of selective collaboration because, during server aggregation, local GNNs intensify their biased patterns rather than assimilate complementary knowledge.
To counter this, we introduce \mypapertitle{}, which combines automatic CL on each client with robust weighted model aggregation on the server.
The approach steers local GNNs toward gradual personalization and shifts server aggregation from exchanging generic knowledge to prioritizing client-specific insights.
Experimental results demonstrate that \mypapertitle{} generally outperforms existing FL frameworks, showing outstanding performance across various benchmark datasets.
Further analyses emphasize the impact of CL on server aggregation, the suitability of our adaptive CL strategy for personalized Subgraph FL, and the proper intensity of collaboration.

\section*{Acknowledgements}
This work was supported by the Institute of Information \& Communications Technology Planning and Evaluation (IITP) and the National IT Industry Promotion Agency (NIPA) through grants funded by the Korean government (MSIT) (RS-2019-II190421, IITP-2025-RS-2020-II201821, RS-2025-02218768, RS-2025-25443718, RS-2025-25442569, RS-2025-02653113, H0601-24-1023). ChatGPT is used for \textbf{writing} (English refinement, typo correction) and \textbf{code} (bug-fix suggestions); all AI outputs are reviewed and edited by the authors. 

\clearpage
\balance
\printbibliography
\clearpage
\nobalance
\appendix




\section{Server Aggregation Stage Algorithm}

After the local training stage, each client uploads the parameters of the local Graph Neural Network (GNN) to the server.
In Algorithm \ref{7:alg:server}, the set $\mathbb{S}$ containing the mask matrices of each local GNN for the random graph is initialized (Line 1).
The central server optimizes the mask matrix for all local GNNs on the random graph (Lines 2-6).
Subsequently, the local GNN parameters are aggregated based on Equations \eqref{4:eq:refinedSimilarity} and \ref{4-1:eq:aggregation} (Lines 7-8).
Note that $\Tilde{\mathbf{S}}^{(t)}_k$ can be represented as a vector of length $|\Tilde{\mathcal{E}}|$ where $|\Tilde{\mathcal{E}}|$ is number of edges in random graph $\Tilde{G}$.
This vectorization offers a more efficient approach to determining client similarity rather than operating on matrices.

\begin{algorithm}[t]
\caption{Server Aggregation Stage for client $k$}
\begin{flushleft}
\textbf{Input}: Total number of clients $K$, set of locally trained GNN parameters for all clients $\{\Bar{\textbf{W}}_i^{(t)}\}_{i=1}^{K}$, random graph $\Tilde{G}$, and scaling factor $\tau$

\textbf{Output}: Aggregated local GNN parameters $\mathbf{W}_k^{(t+1)}$
\end{flushleft}

\begin{algorithmic}[1] 
\STATE Initialize set $\mathbb{S} \gets \emptyset$
\FOR{each client $i$ from 1 to $K$}
    \STATE Optimize $\Tilde{\mathbf{S}}^{(t)}_i$ according to Equation \eqref{4:eq:clloss}
    \STATE $\Tilde{\mathbf{S}}^{(t)}_i \leftarrow \text{CLIP}(\Tilde{\mathbf{S}}^{(t)}_i)$
    \STATE Add element $\Tilde{\mathbf{S}}^{(t)}_i$ to $\mathbb{S}$
\ENDFOR
\vspace{1mm}
\STATE Using the $\mathbb{S}$, compute \\ \quad \quad \quad \quad \quad $\mathbf{W}_k^{(t+1)} \leftarrow \sum_{n=1}^{K} \frac{\exp(\tau \cdot \operatorname{Sim}(k,n))}{\sum_p \exp(\tau \cdot \operatorname{Sim}(k,p))} \Bar{\mathbf{W}}_{n}^{(t)}$

\end{algorithmic}
\label{7:alg:server}
\end{algorithm}

\section{Detailed Theoretical Analysis}
\label{sec:theory}

We provide a theoretical analysis of (i) the robustness of \mypapertitle{}'s client similarity estimation method and (ii) how \emph{Incremental Edge Selection} (IES) enhances overfitting mitigation in federated learning over subgraphs.
Section~\ref{sec:theory:prelims} introduces key assumptions and notation. Section~\ref{sec:theory:community} presents a \emph{Cluster Preservation} theorem with a formal proof, and Section~\ref{sec:theory:generalization} provides a \emph{Generalization Bound} theorem that explicitly leverages uniform convergence arguments.

\subsection{Preliminaries and Assumptions}
\label{sec:theory:prelims}

\begin{assumption}[Node Embeddings.]
Each client $k$ in a federated system is associated with a local subgraph $G_k = (\mathcal{V}_k,\mathcal{E}_k)$. Let $\mathbf{h}_{k,i}^{(t)} \in \mathbb{R}^d$ denote the embedding of node $i \in \mathcal{V}_k$ at round $t$ (the dimension $d$ is typically the GNN’s output dimension). A \emph{reconstructability function} $\kappa: \mathbb{R}^d \times \mathbb{R}^d \to \mathbb{R}$ maps these embeddings to a scalar:
\[
    r_k^{(t)}(i,j)
    \;=\;
    \kappa\!\bigl(\mathbf{h}_{k,i}^{(t)}, \mathbf{h}_{k,j}^{(t)}\bigr),
\]
which represents the likelihood of an edge $(i,j)$ from the perspective of client $k$’s GNN. We assume $\kappa$ is $L$-Lipschitz in each argument:
\end{assumption}

\begin{assumption}[Clusters and Coherence.]
We suppose there are $M$ latent clusters $C_1,\dots,C_M$, each grouping clients with similar data properties, and each client $k$ is primarily associated with one of them, denoted $C(k)$.
If $k$ and $m$ share the same cluster, they produce \emph{similar embeddings} on a reference node set $\mathcal{V}_\mathrm{rand}$:
\begin{equation}
\label{eq:coherence}
    \mathbb{E}_{i \in \mathcal{V}_\mathrm{rand}}
    \|\mathbf{h}_{k,i}^{(t)}
      - \mathbf{h}_{m,i}^{(t)}\|_2
    \;\le\;
    \epsilon_C,
\end{equation}
for some small $\epsilon_C>0$. Clients in different clusters can exceed this limit, ensuring a separation in their node embeddings.
\end{assumption}

\begin{assumption}[Easy vs.\ Hard Edges.]
Each client $k$ partitions its local edge set $\mathcal{E}_k$ into \emph{easy} edges $\mathcal{E}_k^e$ and \emph{hard} edges $\mathcal{E}_k^h$, such that at round $t$,
\[
    \min_{(i,j)\in\mathcal{E}_k^e} r_k^{(t)}(i,j)
    \;\ge\;
    \max_{(i,j)\in\mathcal{E}_k^h} r_k^{(t)}(i,j).
\]
\emph{Incremental Edge Selection} (IES) retains all easy edges from the outset and gradually incorporates a fraction of the hard edges, thus restricting local subgraph complexity early in training.
\end{assumption}

\subsection{Cluster Preservation}
\label{sec:theory:community}

Let $\tilde{\mathbf{A}}_k^{(t)} \in \mathbb{R}^{|\tilde{\mathcal{V}}| \times |\tilde{\mathcal{V}}|}$ be the \emph{reconstructed} adjacency matrix of client $k$ at round $t$, whose $(i,j)$ entry is $r_k^{(t)}(i,j)$, derived from the same graph among all clients.
We prove that same-cluster pairs are strictly closer in Frobenius norm than cross-cluster pairs.

\noindent \paragraph{(Restated) \textsc{Theorem \ref{thm:community}}}(\textsc{Cluster Preservation}) \textit{Suppose clients $k$ and $m$ belong to the same cluster $C$ of similar data properties, whereas client $n$ does not. Let $\tilde{\mathbf{A}}_k^{(t)}$, $\tilde{\mathbf{A}}_m^{(t)}$, and $\tilde{\mathbf{A}}_n^{(t)}$ be their reconstructed adjacency matrices at round $t$ for the same input graph. Under the Lipschitz property of $\kappa$ and the coherence assumption in Equation \eqref{eq:coherence}, there exists $\xi>0$ such that}
\begin{equation}
\label{eq:community-preserve-re}
    \|\tilde{\mathbf{A}}_k^{(t)} 
      - \tilde{\mathbf{A}}_m^{(t)}\|_F
    \;\;\le\;\;
    \frac{1}{\xi}
    \,\bigl\|\tilde{\mathbf{A}}_k^{(t)}
            - \tilde{\mathbf{A}}_n^{(t)}\bigr\|_F
    \;+\;
    \mathcal{O}\!\bigl(\epsilon_C\bigr),
\end{equation}
\textit{for some small $\epsilon_C>0$. Hence, same-cluster reconstructions remain closer than cross-cluster reconstructions.}


\vspace{-2mm}
\begin{proof}
We analyze the squared Frobenius norm of the difference in reconstructability:
\[
    \|\tilde{\mathbf{A}}_p^{(t)}
      - \tilde{\mathbf{A}}_q^{(t)}\|_F^2
    \;=\;
    \sum_{(i,j)}
    \Bigl(r_p^{(t)}(i,j) - r_q^{(t)}(i,j)\Bigr)^2.
\]
Suppose clients $p$ and $q$ lie in the same cluster, so by \eqref{eq:coherence}, we have 
\(\mathbb{E}_{i\in\mathcal{V}_\mathrm{rand}}\|\mathbf{h}_{p,i}^{(t)} - \mathbf{h}_{q,i}^{(t)}\|\le \epsilon_C.\)
Since $r_p^{(t)}(i,j) = \kappa(\mathbf{h}_{p,i}^{(t)}, \mathbf{h}_{p,j}^{(t)})$ and $\kappa$ is $L$-Lipschitz, it follows that
\[
    \bigl|r_p^{(t)}(i,j) - r_q^{(t)}(i,j)\bigr|
    \;\le\;
    L\,\|\mathbf{h}_{p,i}^{(t)}-\mathbf{h}_{q,i}^{(t)}\|
    + L\,\|\mathbf{h}_{p,j}^{(t)}-\mathbf{h}_{q,j}^{(t)}\|.
\]
Averaging this bound over nodes $i,j$ in $\mathcal{V}_\mathrm{rand}$ yields a difference of order $O(\epsilon_C)$.  Summing across all $(i,j)$ then shows
\[
    \|\tilde{\mathbf{A}}_p^{(t)}
      - \tilde{\mathbf{A}}_q^{(t)}\|_F
    \;\le\;
    \mathcal{O}(\epsilon_C).
\]
In contrast, if clients $p$ and $q$ belong to different clusters, their node embedding distributions diverge more substantially, leading to a typical difference of at least $\xi \,\epsilon_C$ for some $\xi>1$. Thus, the cross-cluster difference 
\(\|\tilde{\mathbf{A}}_p^{(t)} - \tilde{\mathbf{A}}_q^{(t)}\|_F\)
becomes $\Omega(\xi\,\epsilon_C)$. Combining these yields the ratio implied by \eqref{eq:community-preserve-re}, namely that
\[
    \|\tilde{\mathbf{A}}_k^{(t)} 
      - \tilde{\mathbf{A}}_m^{(t)}\|_F
    \;\le\;
    \tfrac{1}{\xi}
    \,\|\tilde{\mathbf{A}}_k^{(t)}
         - \tilde{\mathbf{A}}_n^{(t)}\|_F
    \;+\;
    \mathcal{O}(\epsilon_C),
\]
\end{proof}
Equation \eqref{eq:community-preserve-re} shows that clients in the same cluster produce near-identical reconstructed adjacency matrices; this sharp contrast with cross-cluster reconstructions—revealed by the shared random graph—lets the server reliably infer clusters during training.

\subsection{Generalization Bound and Overfitting Control}
\label{sec:theory:generalization}

We prove how IES prevents overfitting by restricting the fraction of harder edges, thereby controlling the complexity of each client’s local training.
Define $\mathcal{L}_k^{(t)}$ as the training loss of client $k$ at round $t$ (empirically measured on the retained edges) and $\mathcal{L}_k^{\text{test}}$ as the expected test loss under the true data distribution. Let $|\mathcal{E}_k^e|$ be the number of easy edges and $|\mathcal{E}_k^h|$ the number of hard edges in $\mathcal{E}_k$.

\begin{figure}[t]
    \centering
    \includegraphics[width=0.95\columnwidth]{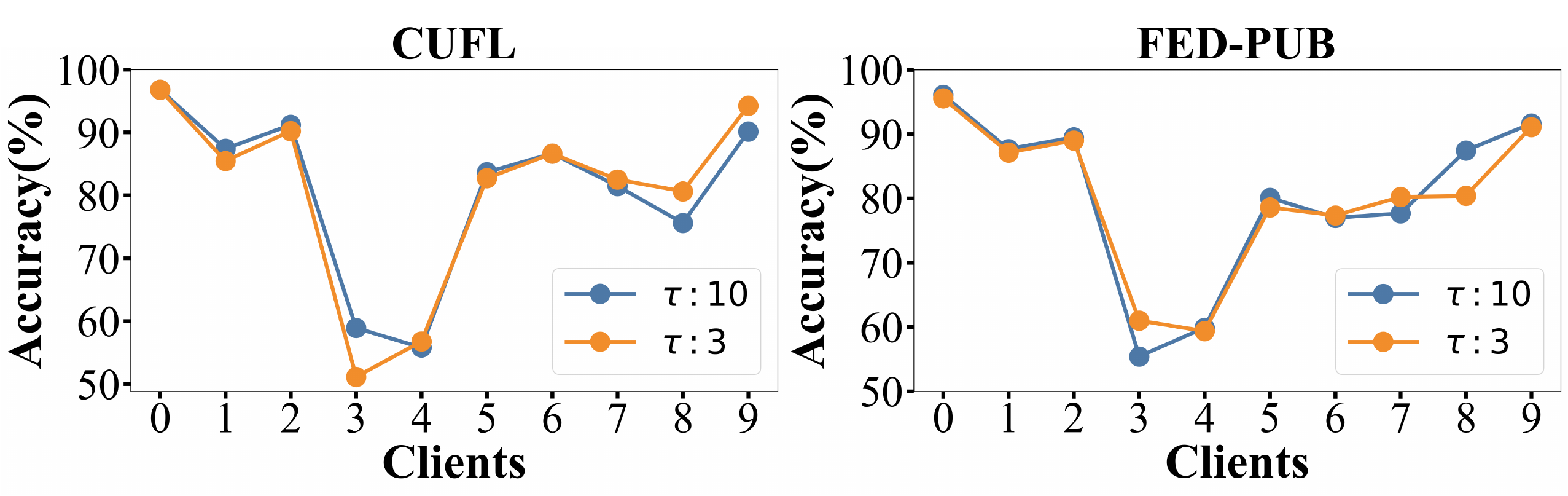}
    \caption{Performance per client according to $\tau$ on Cora.}
    \Description{Tau Motivation}
    \label{7:fig:tauMotivation}
\end{figure}

\begin{theorem}[Generalization Bound]
\label{thm:gen-bound}
Suppose that at round $t$, client $k$ retains \emph{all} its easy edges $\mathcal{E}_k^e$ and a fraction $0 \le \lambda^{(t)} \le 1$ of its hard edges $\mathcal{E}_k^h$, so the local subgraph used has size at most $|\mathcal{E}_k^e| + \lambda^{(t)}|\mathcal{E}_k^h|$. Then, under standard uniform-convergence arguments (e.g., \ Rademacher complexity or a VC-type bound), there exists a universal constant $c_0>0$ such that, with probability at least $1-\delta$,
\begin{equation}
\label{eq:generalization}
    \mathbb{E}\bigl[\mathcal{L}_k^{\text{test}}\bigr]
    \;\le\;
    \mathcal{L}_k^{(t)}
    \;+\;
    c_0\,\sqrt{\frac{|\mathcal{E}_k^e|
     \;+\; \lambda^{(t)}\,|\mathcal{E}_k^h|}{|\mathcal{E}_k|}}
    \;+\;
    \mathcal{O}\Bigl(\sqrt{\tfrac{\log(1/\delta)}{|\mathcal{E}_k|}}\Bigr).
\end{equation}
Hence, limiting the fraction of hard edges $\lambda^{(t)}$ shrinks the capacity term, reducing the gap between training loss and expected test loss.
\end{theorem}

\begin{proof}
By uniform convergence (see, e.g., \cite{Bartlett2002Rademacher}), the generalization error $\mathcal{L}_k^{\text{test}}(\mathbf{w}) - \mathcal{L}_k^{(t)}(\mathbf{w})$ can be bounded by a function of the hypothesis-class capacity. In adjacency-based GNNs, restricting the local subgraph to $|\mathcal{E}_k^e| + \lambda^{(t)}|\mathcal{E}_k^h|$ edges decreases the number of possible adjacency patterns and thus the effective size of the function class.  Rademacher complexity or VC-bounds then yield
\[
    \mathrm{Complexity}\bigl(\mathbf{w}\bigr)
    \;=\;
    \mathcal{O}\!\Bigl(\sqrt{\tfrac{|\mathcal{E}_k^e| 
      + \lambda^{(t)}|\mathcal{E}_k^h|}{|\mathcal{E}_k|}}\Bigr)
    + \mathcal{O}\Bigl(\sqrt{\tfrac{\log(1/\delta)}{|\mathcal{E}_k|}}\,\Bigr).
\]
Adding $\mathcal{L}_k^{(t)}$ to both sides of this capacity term implies \eqref{eq:generalization}, completing the proof.
\end{proof}

Inequality~\eqref{eq:generalization} states that the test loss is \emph{upper-bounded} by the training loss plus a function of (i) the fraction $\lambda^{(t)}$ of hard edges and (ii) the usual $\sqrt{\log(1/\delta)/|\mathcal{E}_k|}$ deviation term. As $\lambda^{(t)}$ remains small in early rounds, local overfitting is greatly mitigated. Once the GNN stabilizes through federated updates, $\lambda^{(t)}$ may grow without incurring excessive variance.

\section{Adaptive \texorpdfstring{$\tau$}{tau} Scheduler
\label{appen:ATS}}

In this section, we introduce our adaptive $\tau$ scheduler, briefly mentioned in Section \ref{expr:VSF}.
Similar to the GreedyLR scheduler \cite{subramanian2023zeroth}, the adaptive $\tau$ scheduler greedily updates the personalized $\tau$ for each client at the beginning of specific training rounds. 
The main intuition here is that the optimal $\tau$ varies across clients in personalized Subgraph FL employing weighted model aggregation, as empirically demonstrated in Figure \ref{7:fig:tauMotivation}.
More concretely, while the higher $\tau$ tends to be preferred as the data heterogeneity gets severe, the $\tau$ value that yields optimal performance for the local GNN varies from client to client (See Table \ref{5:table:tau}).

\setlength{\textfloatsep}{2mm}
\setlength{\floatsep}{2mm}

\begin{algorithm}[t]
\caption{Adaptive $\tau$ scheduler}

\begin{flushleft}
\textit{// We omit the client's index $k$ in the notation. }

\textbf{Input}: Scaling factor of previous round is $\tau^{(t-1)}$, performance from the previous round $p^{(t-1)}$, performance from the current round $p^{(t)}$, update patience $patience$, update direction $s$, number of consecutive improved rounds $r_{good}$, number of consecutive deteriorated rounds $r_{bad}$, multiplicative factor $\rho_\tau$, minimum scaling factor $\tau_{min}$, and maximum scaling factor $\tau_{max}$

\textbf{Output}: Scaling factor of current round $\tau^{(t)}$
\end{flushleft}

\begin{algorithmic}[1] 

\IF{$p^{(t)} \geq p^{(t-1)}$}
    \STATE $r_{good} \leftarrow r_{good} + 1$
    \STATE $r_{bad} \leftarrow 0$
\ELSE
    \STATE $r_{bad} \leftarrow r_{bad} + 1$
    \STATE $r_{good} \leftarrow 0$
\ENDIF

\IF{$r_{good} > patience$}
    \STATE $\tau'^{(t)} \leftarrow \rho_\tau^s\cdot\tau^{(t-1)}$
    \STATE $r_{good} \leftarrow 0$
\ELSIF{$r_{bad} > patience$}
    \STATE $s \leftarrow (-1) \cdot s$
    \STATE $\tau'^{(t)} \leftarrow \rho_\tau^s\cdot\tau^{(t-1)}$
    \STATE $r_{bad} \leftarrow 0$
\ELSE
    \STATE $\tau'^{(t)} \leftarrow \tau^{(t-1)}$
\ENDIF

\STATE $\tau^{(t)} \leftarrow
\begin{cases} 
\tau_{min} & \text{if } \tau'^{(t)} \leq \tau_{min}, \\
\tau'^{(t)} & \text{if } \tau_{min} < \tau'^{(t)} \leq \tau_{max}, \\
\tau_{max} & \text{if } \tau'^{(t)} > \tau_{max}.
\end{cases}$

\end{algorithmic}
\label{7:alg:tau}
\end{algorithm}

The adaptive $\tau$ is outlined as follows.
In the first round, the adaptive $\tau$ scheduler initializes $\tau$ to 5, and $r_{good}$ and $r_{bad}$ to 0, which denote the number of consecutive improved and deteriorated rounds, respectively.
The parameter $\eta \in \{-1, 1\}$, which is initialized to 1, determines the direction of the update.
If the number of consecutive performance changes, whether $r_{good}$ or $r_{bad}$, surpass the $patience$ set to 5, $\tau$ is adjusted by scaling it with an update step size $\xi$, which is fixed at 1.25, depending on the sign of $\eta$.
The value of $\tau$ is clipped to $\tau_{max}=10$ when it exceeds the upper bound, or to $\tau_{min}=3$ when it falls below this minimum.
The update interval for $\tau$ is set to either 1 or 5.


\setlength{\textfloatsep}{\origtextfloatsep}
\setlength{\floatsep}{\origfloatsep}

\begin{table*}[t]
\caption{Dataset statistics. The number of nodes, edges, and classes for each setting is described. ``Full" refers to the original global network.}
\setlength\extrarowheight{1pt}
\centering
\resizebox{\textwidth}{!}{
\begin{tabular}{c cccc cccc cccc}
\noalign{\vskip 0.5mm} \Xhline{2\arrayrulewidth} \noalign{\vskip 0.5mm}
\multicolumn{1}{c}{} & \multicolumn{4}{c}{\textbf{Cora}} & \multicolumn{4}{c}{\textbf{CiteSeer}} & \multicolumn{4}{c}{\textbf{PubMed}} \\ 
\cmidrule(lr){2-5}\cmidrule(lr){6-9}\cmidrule(lr){10-13}
& \textbf{Full}     & \textbf{5 Clients}       & \textbf{10 Clients}     & \textbf{20 Clients}    & \textbf{Full}    & \textbf{5 Clients}      & \textbf{10 Clients}    & \textbf{20 Clients}    & \textbf{Full}      & \textbf{5 Clients}       & \textbf{10 Clients}     & \textbf{20 Clients}     \\ \hline \noalign{\vskip 0.5mm}
\multicolumn{1}{c|}{\# Classes} & \multicolumn{4}{c}{7}               & \multicolumn{4}{c}{6}            & \multicolumn{4}{c}{3}                 \\
\multicolumn{1}{c|}{\# Nodes}   & 2,485    & 497     & 248    & 124   & 2,120   & 424    & 212   & 106   & 19,717    & 3,943   & 1,971  & 985    \\
\multicolumn{1}{c|}{\# Edges}   & 5,429    & 933     & 445    & 210   & 3,679   & 704    & 337   & 162   & 44,338    & 8,187   & 3,835  & 1,803  \\  \noalign{\vskip 0.5mm} \hline

\noalign{\vskip 0.5mm}\noalign{\vskip 0.5mm}
\multicolumn{1}{c}{} & \multicolumn{4}{c}{\textbf{Amazon-Computer}} & \multicolumn{4}{c}{\textbf{Amazon-Photo}} & \multicolumn{4}{c}{\textbf{ogbn-arxiv}} \\ 
\cmidrule(lr){2-5}\cmidrule(lr){6-9}\cmidrule(lr){10-13}
& \textbf{Full}     & \textbf{5 Clients}       & \textbf{10 Clients}     & \textbf{20 Clients}    & \textbf{Full}    & \textbf{5 Clients}      & \textbf{10 Clients}    & \textbf{20 Clients}    & \textbf{Full}      & \textbf{5 Clients}       & \textbf{10 Clients}     & \textbf{20 Clients}     \\ \hline 
\noalign{\vskip 0.5mm}
\multicolumn{1}{c|}{\# Classes} & \multicolumn{4}{c}{10}              & \multicolumn{4}{c}{8}            & \multicolumn{4}{c}{40}                \\
\multicolumn{1}{c|}{\# Nodes}   & 13,381   & 2,676   & 1,338  & 669   & 7,487   & 1,497  & 748   & 374   & 169,343   & 33,868  & 16,934 & 8,467  \\
\multicolumn{1}{c|}{\# Edges}   & 245,861  & 42,240  & 18,067 & 7,816 & 119,043 & 21,569 & 9,660 & 4,273 & 1,166,243 & 205,474 & 91,112 & 43,377 \\  \noalign{\vskip 0.5mm} \hline
\end{tabular}
}
\label{7:table:dataset}
\vspace{-3mm}
\end{table*}

\begin{figure*}[!t]
    \centering
    \includegraphics[width=0.92\textwidth]{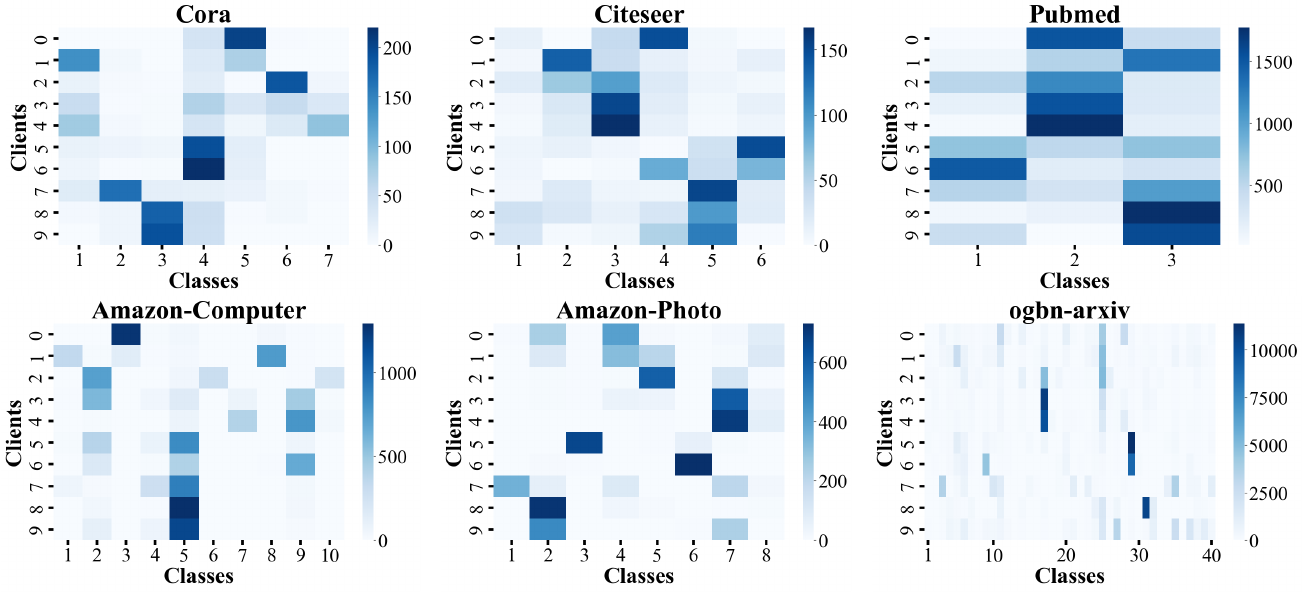}
    \caption{Class distribution of datasets. Darker color indicates that more nodes belong to the corresponding class.}
    \label{7:fig:datasetDesc}
    \Description{Dataset Description}
\end{figure*}

In Algorithm \ref{7:alg:tau}, successive performance changes are tracked (Lines 1-7).
The adaptive $\tau$ scheduler measures the client's performance changes based on the accuracy of the aggregated local GNN.
If the number of continuous performance improvements exceeds the specified $patience$, $\tau$ is adjusted in the current direction of the update (Lines 8-10).
Conversely, if the number of continuous performance declines goes beyond the $patience$, the update direction $\eta$ is reversed, and $\tau$ is subsequently refined (Lines 11-14).
$\tau$ is not altered if the performance variation stays within the $patience$ (Lines 15-16).
In the final step, $\tau$ is constrained to lie within the prescribed range (Line 18).
By greedily modifying $\tau$ based on performance trends, the adaptive $\tau$ scheduler calibrates the intensity of collaboration within the community, steering the model towards enhanced performance.
Consequently, the dynamic update of $\tau$ effectively elevates the performance of personalized FL that utilizes the similarity matrix.

\section{Experimental Setups}

\subsection{Dataset Descriptions}

Table \ref{7:table:dataset} presents statistics of the six datasets used in our experiments.
These include small citation graphs such as Cora, CiteSeer, and PubMed \cite{DBLP:conf/icml/YangCS16}; product graphs such as Amazon-Computer and Amazon-Photo \cite{DBLP:journals/corr/abs-1811-05868}; and a large citation graph, ogbn-arxiv \cite{DBLP:conf/nips/HuFZDRLCL20}.
In our main experimental setting, the local subgraphs are the output of METIS \cite{karypis1997metis}, ensuring no overlapping nodes between them.
Due to the missing links between local subgraphs, the total number of edges across clients is lower than the number of edges in the full graph.
Furthermore, the homophily principle \cite{mcpherson2001birds}, which states that nodes with the same label are more likely to be connected, leads to heterogeneity in class distribution (See Figure \ref{7:fig:datasetDesc}).



\subsection{Baselines}

\subsubsection{FedAvgCL}

FedAvgCL is a centralized Subgraph FL baseline developed in this work, which applies Incremental Edge Selection (IES) to FedAvg \cite{DBLP:conf/aistats/McMahanMRHA17}.
During the local training stage, local GNN is optimized following the curriculum provided by IES.
In the aggregation stage, the server aggregates the local GNNs with respect to the number of nodes in each local subgraph.
FedAvgCL demonstrates the necessity of adopting Curriculum Learning (CL) in Subgraph FL and the importance of weighted model aggregation in \mypapertitle{}.

\subsubsection{FedGNN}

FedGNN \cite{DBLP:journals/corr/abs-2102-04925} augments local subgraphs using overlapping nodes as expansion pivots.
This design, however, is not applicable to our experimental setting, where client subgraphs are strictly disjoint.
For fair comparison, we adapt FedGNN to the node non-overlapping scenario.
Concretely, we sample 10\% of nodes from each client to construct a shared candidate pool, which serves as a surrogate for overlaps.
Each local node subsequently augments its neighborhood with candidates chosen based on cosine similarity in the embedding space of the global model.
The number of attached neighbors for each node is 20 for Cora and CiteSeer, 10 for PubMed, Amazon-Computer, and Amazon-Photo, and 1 for ogbn-arxiv.

\subsection{Additional Ablation Studies}

\subsubsection{Results on Louvain Partitioning}

To examine the robustness of \mypapertitle{} under alternative subgraph partitioning schemes, we apply \mypapertitle{} to subgraphs generated by the Louvain algorithm \cite{DBLP:journals/corr/abs-2311-06047}.
The Louvain method greedily maximizes modularity in community detection, thereby favoring structurally homogeneous clusters. 
As a result, structurally similar clients become more pronounced, but the resulting subgraph sizes are highly imbalanced.
Since the Louvain method inherently leaves the number of partitions unspecified, we adopt the strategy of \cite{DBLP:conf/nips/ZhangYLSY21} to randomly merge the resulting subgraphs irrespective of graph properties.
Table~\ref{tab:louvain_performance} demonstrates that \mypapertitle{} achieves strong performance on Louvain-partitioned subgraphs. 
The variance across runs remains limited, further suggesting that \mypapertitle{} exhibits stable behavior under this partitioning.

\subsubsection{Influence of Random Graph Configurations}

\begin{table}[t]
\caption[Performance on the Louvain Graph Partitioning]{Performance on Louvain-partitioned subgraphs.}
\vspace{-2mm}
\begin{center}
\begin{small}
\resizebox{\columnwidth}{!}{%
\begin{tabular}{clllll}
\Xhline{2\arrayrulewidth}
                            & \multicolumn{2}{c}{\textbf{Cora}} 
                            & \multicolumn{2}{c}{\textbf{Amazon-Photo}} \\ \cline{2-3} \cline{4-5}
\textbf{Frameworks}         & \textbf{10 Clients} & \textbf{20 Clients}
                            & \textbf{10 Clients} & \textbf{20 Clients} \\ \hline
FedProx                     & 76.93$\pm$0.42 & 68.85$\pm$1.27
                            & 91.47$\pm$0.41 & 89.13$\pm$1.05 \\
FED-PUB                     & \underline{81.29$\pm$0.23} & \underline{77.79$\pm$0.39}
                            & \underline{94.15$\pm$0.21} & \underline{93.20$\pm$0.15} \\ \hline
\mypapertitle{} (Ours)                        & \textbf{82.79$\pm$0.10} & \textbf{79.80$\pm$0.30}
                            & \textbf{95.40$\pm$0.10} & \textbf{95.11$\pm$0.08} \\ 
\Xhline{2\arrayrulewidth}
\end{tabular}}
\end{small}
\end{center}
\label{tab:louvain_performance}
\vspace{-2mm}
\end{table}


\begin{table}[t]
\caption[Performance with Different Random Graph Generators]{Performance on reference graphs from different random graph models.}
\vspace{-2mm}
\begin{center}
\begin{small}
\resizebox{\columnwidth}{!}{%
\begin{tabular}{cllll}
\Xhline{2\arrayrulewidth}
                                & \multicolumn{2}{c}{\textbf{Cora}} & \multicolumn{2}{c}{\textbf{PubMed}} \\ \cline{2-5}
\textbf{Graph Generator}        & \textbf{10 Clients} & \textbf{20 Clients} & \textbf{10 Clients} & \textbf{20 Clients} \\ \hline
Erdős–Rényi                     & 82.10$\pm$0.25 & 78.76$\pm$0.56 & 84.36$\pm$0.04 & 84.50$\pm$0.29 \\
Barabási–Albert                 & \textbf{84.16$\pm$0.18} & \underline{79.48$\pm$0.93} & \underline{85.56$\pm$0.05} & \textbf{86.31$\pm$0.23} \\
SBM                             & \underline{83.98$\pm$0.19} & \textbf{79.62$\pm$0.23} & \textbf{86.22$\pm$0.02} & \underline{86.17$\pm$0.18} \\
\Xhline{2\arrayrulewidth}
\end{tabular}}
\end{small}
\end{center}
\vspace{-2mm}
\label{tab:random_graph_generators}
\end{table}

\begin{table}[t]
\caption[Performance with Different Graph Size]{Performance on reference graphs with different random graph sizes.}
\vspace{-2mm}
\begin{center}
\begin{small}
\resizebox{\columnwidth}{!}{%
\begin{tabular}{cllll}
\Xhline{2\arrayrulewidth}
                                & \multicolumn{2}{c}{\textbf{Cora}} & \multicolumn{2}{c}{\textbf{PubMed}} \\ \cline{2-5}
\textbf{Graph Size}        & \textbf{10 Clients} & \textbf{20 Clients} & \textbf{10 Clients} & \textbf{20 Clients} \\ \hline \rule{0pt}{2.5ex}
$|\tilde{V}|=100$                     & 83.61$\pm$0.33 & 79.25$\pm$0.42 & 85.93$\pm$0.06 & 86.03$\pm$0.23 \\
$|\tilde{V}|=500$                 & \textbf{83.98$\pm$0.19} & \textbf{79.62$\pm$0.23} & \textbf{86.22$\pm$0.02} & \underline{86.17$\pm$0.18} \\
$|\tilde{V}|=1000$                             & \underline{83.96$\pm$0.12} & \underline{79.41$\pm$0.22} & \underline{86.18$\pm$0.03} & \textbf{86.21$\pm$0.16} \\
\Xhline{2\arrayrulewidth}
\end{tabular}}
\end{small}
\end{center}
\vspace{-4mm}
\label{tab:random_graph_size}
\end{table}

The random graphs allow \mypapertitle{} to estimate client similarity without exposing private data, yet our experiments rely solely on the Stochastic Block Model (SBM) with a fixed graph size.
In this section, we examine how random graph selection affects performance.
Accordingly, we vary the random graph model and size.

Regarding the random graph model, we substitute SBM with two widely studied alternatives---Erdős–Rényi (ER) and Barabási–Albert (BA) \cite{DBLP:journals/corr/abs-2403-14415}, while keeping the number of nodes consistent with the hyperparameter setting.
ER draws each edge independently with a fixed probability, producing a nearly binomial degree distribution and virtually no intrinsic community structure, whereas BA grows the network by preferential attachment, yielding a heavy-tailed degree distribution dominated by hub nodes.
Table~\ref{tab:random_graph_generators} summarizes the performance on Cora and PubMed, showing that both SBM and BA consistently outperform ER.
SBM and BA encode salient citation-network patterns---communities and hub-dominated connectivity---whereas ER remains structure-agnostic.
This indicates that \mypapertitle{} performs better when the random graphs encode the clients' structural statistics.


Orthogonal to the choice of random graph model, we examine the impact of graph size by varying the number of nodes.
All other settings are kept consistent with the original hyperparameters.
As shown in Table~\ref{tab:random_graph_size}, undersized random graphs fail to capture representative features of local GNNs.
Increasing the size provides more structural information, which initially improves performance.
However, once the size reaches a large scale, improvements become marginal and, in some cases, larger graphs impair accuracy.
While larger random graphs can improve performance, their size simultaneously increases the complexity of the server aggregation stage.
Therefore, before the random graph size becomes excessively large, a trade-off emerges between performance and complexity.
\clearpage

\end{document}